%% file: main_rlhf.tex
\documentclass{fairmeta}

\usepackage[T1]{fontenc}
\usepackage{graphicx} 
\usepackage{multirow}
\usepackage{array}
\usepackage{booktabs}
\usepackage{adjustbox}
\usepackage{graphicx}
\usepackage{subcaption}
\usepackage{epigraph}
\usepackage{tcolorbox}
\definecolor{RoyalBlue}{RGB}{65,105,225}
\definecolor{Green}{RGB}{34,139,34}
\hypersetup{
    colorlinks=true,
    citecolor=RoyalBlue,
    linkcolor=RoyalBlue,
    filecolor=magenta,      
    urlcolor=RoyalBlue
}
\usepackage{multicol}
\usepackage{graphicx,wrapfig}
\usepackage{enumitem}
\usepackage{amsmath,amsbsy,amsfonts,amssymb,amsthm}
\usepackage{bm}
\usepackage{newtx}
\usepackage{algorithm,algorithmic,mathtools}
\usepackage{color,cases}
\usepackage{tikz,bbm}
\usepackage{dsfont}
\usetikzlibrary{calc,shapes}
\usepackage{hyperref,comment}
\usepackage{xcolor,url,verbatim}
\usepackage{afterpage}
\DeclareMathOperator{\Supp}{Supp}

\DeclareMathOperator{\dist}{dist}

\DeclareMathOperator{\poly}{poly}

\newcommand{\abs}[1]{\left\lvert #1 \right\rvert}
\newcommand{\paran}[1]{\left( #1 \right)}

\newcommand{\zeronorm}[1]{\left\lVert #1 \right\rVert_{0}}

\newcommand{\twonorm}[1]{\left\lVert #1 \right\rVert_{2}}

\newcommand{\frobnorm}[1]{\left\lVert #1 \right\rVert_{F}}
\newcommand{\infnorm}[1]{\left\lVert #1 \right\rVert_{\infty}}
\newcommand{\nn}{\nonumber}

\newtheorem{assumption}{Assumption}
\newtheorem{claim}{Claim}
\newtheorem{lemma}{Lemma}
\newtheorem{theorem}{Theorem}
\newtheorem{remark}{Remark}
\newtheorem{corollary}{Corollary}

\newcommand{\se}{\widehat{\sigma}}
\newcommand{\pe}{\widehat{P}}
\newcommand{\pc}{P^c}
\newcommand{\tn}{\widetilde{N}}
\newcommand{\ts}{\widetilde{S}}
\newcommand{\tp}{\widetilde{P}}

\newcommand{\pec}{\overline{P}}

\newcommand{\R}{\mathbb{R}}

\renewcommand{\P}{\mathcal{P}}
\renewcommand{\S}{\mathcal{S}}

\title{
\begin{center}
CURATRON: \underline{C}omplete and Rob\underline{u}st P\underline{r}eference D\underline{at}a for \underline{R}ig\underline{o}rous Alig\underline{n}ment of Large Language Models
\end{center}}

\author[1]{Son The Nguyen} 
\author[2]{Niranjan Uma Naresh}
\author[1]{Theja Tulabandhula}
\affiliation[1]{University of Illinois Chicago}
\affiliation[2]{Independent Researcher}

\abstract{
This paper addresses the challenges of aligning large language models (LLMs) with human values via preference learning (PL), focusing on incomplete and corrupted data in preference datasets. We propose a novel method for robustly and completely recalibrating values within these datasets to enhance LLMs' resilience against these issues. In particular, we devise a guaranteed polynomial time ranking algorithm that robustifies several existing models, such as the classic Bradley--Terry--Luce (BTL)~\citep{bradley1952rank} model and certain generalizations of it. To the best of our knowledge, our present work is the first to propose an algorithm that provably recovers an $\epsilon$-optimal ranking with high probability while allowing as large as $O(n)$ perturbed pairwise comparison results per model response. Furthermore, we show robust recovery results in the partially observed setting. Our experiments confirm that our algorithms handle adversarial noise and unobserved comparisons well in both general and LLM preference dataset settings. This work contributes to the development and scaling of more reliable and ethically aligned AI models by equipping the dataset curation pipeline with the ability to handle missing and maliciously manipulated inputs.
}

\correspondence{\email{snguye65@uic.edu}, \email{theja@uic.edu}, \email{un.niranjan@gmail.com}}
\begin{document}

\maketitle

\begin{sloppypar}
\input{intro}

\input{model}

\input{analysis}

\input{expt}

\input{conc}

\newpage
\bibliography{refs_ranking}
\bibliographystyle{apalike}
\newpage

\appendix
\input{appendix}
\end{sloppypar}
\end{document}

%% file: intro.tex
\section{Introduction}\label{sec:intro}



Large Language Models (LLMs) are highly advanced Artificial Intelligence (AI) systems capable of understanding, interpreting, and generating languages. The integration of AI chatbots like ChatGPT into our daily lives and businesses has profoundly impacted society and various industries \citep{eloundou2023gpts, brynjolfsson2023generative}. These models have evolved from being specialized tools in specific fields to versatile assets that are increasingly applied in everyday activities and diverse work environments. However, the success of GPTs/LLMs depends not only on their ability to generate responses and perform tasks well but also on their alignment with human values and expectations.

\textbf{\textit{Early challenges in controlling advanced chatbots:}} Before ChatGPT, AI language models and chatbots often produced undesirable answers, making them unsuitable for public use. Examples of this problem include Microsoft's Tay chatbot \citep{Lee_2016}, the National Eating Disorders Association's Tessa \citep{Mccarthy_2023}, and Google's Meena/LaMDA \citep{DBLP:journals/corr/abs-2001-09977}. Although Meena/Lambda's release was initially stopped due to not complying with Google's AI safety and fairness guidelines, the project was later released publicly as Bard due to pressure from ChatGPT and Bing Chat \citep{businessinsiderGoogleEngineers}. Unfortunately, its factual error ended up costing Alphabet \$100 billion dollars in market cap \citep{timeBardChatbot}. Another recent AI incident at the time of writing this paper involved Gemini's image generation AI model, which was found to be biased in favor of certain minority groups, leading Google to shut it down \citep{forbesGooglePauses}. Attempts have been made to address these problems by measuring the personality traits of large language models and modifying them through direct fine-tuning techniques \citep{karra2023estimating}. However, it was not until Instruct-GPT was released with Reinforcement Learning from Human Feedback (RLHF) \citep{ouyang2022training}, that marked a significant breakthrough in aligning LLMs effectively with human values. RLHF has paved the way for OpenAI's ChatGPT to reshape industries, and GPT-4 to be considered a preliminary form of artificial general intelligence (AGI) \citep{bubeck2023sparks}, a type of AI that is vastly more intelligent than humans.

\textbf{\textit{Preference Learning (PL) as a solution:}} Our examples above demonstrate that managing AI has long been recognized as imperative within both businesses and research communities \citep{berente2021managing}. However, both sectors have been struggling with the unpredictable and sometimes adverse behaviors that stem from the increased autonomy and learning capabilities of generative models. Fortunately, alignment through preference learning (PL) offers an effective way to guide these models toward outcomes that align with human values and intentions, much like the approach used to manage human agents. Current PL methods for AI and LLMs are largely achieved through fine-tuning with either human feedback \citep{ouyang2022training} or AI feedback \citep{lee2023rlaif}. This process typically utilizes reinforcement learning algorithms, such as Proximal Policy Optimization (PPO) \citep{schulman2017proximal}, as an indirect fine-tuning approach; or alternatively, a direct fine-tuning approach, such as Direct Preference Optimization (DPO) \citep{rafailov2023direct}. These approaches are usually used in combination with a probabilistic framework, such as Bradley-Terry-Luce, to model pairwise comparisons between model responses and guide the fine-tuning process. Such practical adaptations have been integral to the success of SOTA LLMs such as ChatGPT \citep{ouyang2022training}, Claude \citep{bai2022training}, and Llama \citep{touvron2023llama}, enabling more effective alignment between model outputs and human preferences.

\textbf{\textit{Bottlenecks and Risks of PL:}} Alignment is a crucial step in steering AI models, and preference learning (PL) has proven to be an effective approach; but it entails significant costs, particularly in terms of scaling the collection of human preference data \citep{touvron2023llama2}. The process of acquiring human preferences for LLMs is both time-intensive and costly, requiring substantial resources and the recruitment of qualified human labelers to ensure high-quality data \citep{bai2022training,lee2023rlaif}. For example, training costs for LLaMa-2 were estimated at \$20-45 million, with approximately \$8 million dedicated solely to acquire preference data \citep{llama2details}. Even major AI companies, such as Google, Meta, OpenAI, and Anthropic, face challenges in aligning their models to reflect a wide range of human values and cultural contexts reliably.

One proposed solution to this scaling challenge is the use of automated evaluators, such as LLMs or agents acting as judges, to infer preferences. However, these models also inherit biases from the human data used to train them, making them prone to similar reliability issues and vulnerabilities to manipulation. Furthermore, there is an ongoing debate on whether human involvement should be fully eliminated from the development loop. Another alternative approach is crowd-sourcing the preference data collection process, involving the general public to provide a more diverse set of human values at a lower cost \citep{huang2024collective}. While crowdsourcing can lower costs, it also struggles with data integrity, such as accuracy issues and manipulation risks.

\begin{figure}[h]
    \small
    \centering    \includegraphics[width=0.8\linewidth]{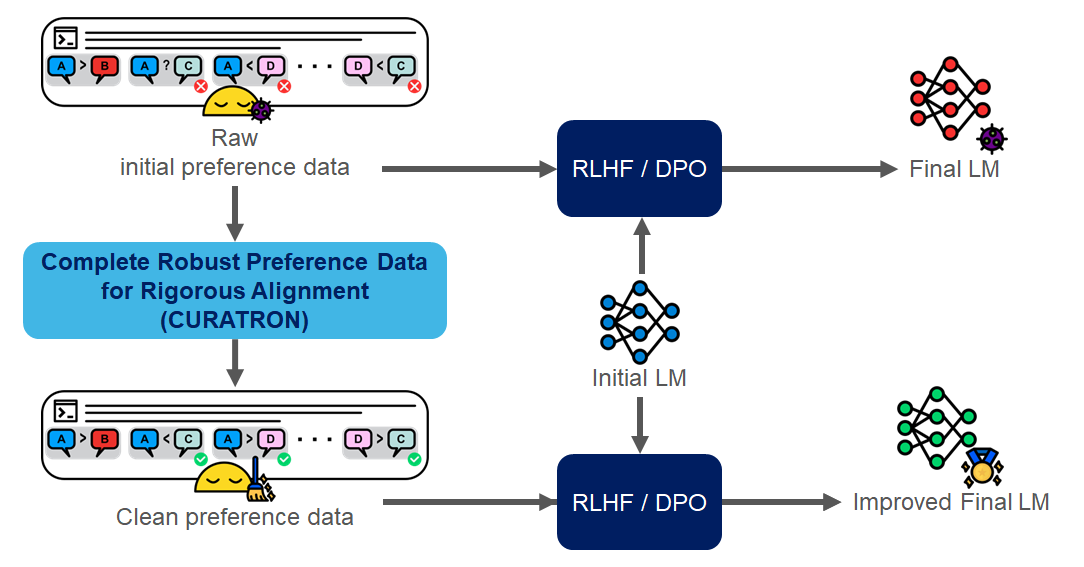}
    \caption{\texttt{CURATRON} corrects incomplete and adversarially corrupted preference data to improve RLHF/DPO alignment results compared to raw initial preference data.}
    \label{fig:systemdiagram}
\end{figure}

From our discussion above, collecting high-quality human preferences is the bottleneck to building better AIs and LLMs. Participants may unintentionally or intentionally provide incomplete, inaccurate, or harmful feedback, as highlighted in \citep{casper2023open,checklistbias,DBLP:journals/corr/abs-1801-02546, COCOS201786}. Using such data directly to train AI can result in suboptimal AI systems that are biased, have misaligned behaviors, or can lead to outcomes that are unfair, discriminatory, or harmful (Figure~\ref{fig:systemdiagram}). 

\setlength{\textfloatsep}{5pt} 
\setlength{\floatsep}{5pt}     

\textbf{\textit{Framing the problem:}} To address the challenges, we frame the following learning problem: Suppose there are $n$ distinct responses we wish to order based on pairwise comparisons with probabilistic outcomes, aimed at training AIs or LLMs to learn human preferences, while having no access to the content of the responses. We are given a full set $\aleph = \{ ( i, j, \{ y_{ij}^k \} ) \}$ consisting of $K$ independent pairwise comparison outcomes, denoted by $\{ y_{ij}^k \} \in \{ 0, 1 \}, k \in [K]$. Between pairs of responses $(i,j) \subseteq [n] \times [n]$, a significant proportion of which might be corrupted by an adversary.
In this passive learning setting, the concrete questions we wish to address are as follows.
\begin{enumerate}[nolistsep,noitemsep]
\item \label{q:id} Is it possible to identify comparisons whose results were corrupted by an adversary?
\item \label{q:rank} Having identified the corrupted results, as desired, is it possible to filter them out while computing a global ranking of the $n$ responses?
\item \label{q:tractable} Is this task statistically and computationally tractable?
\item \label{q:construct} If so, is it possible to construct a provably correct and efficient algorithm, and what are the associated properties?
\item \label{q:expt} Further, does it work well in practice when we may also encounter incomplete or unobserved data?
\end{enumerate}
\textit{\textbf{Our Contributions:}} We systematically answer the above questions in the affirmative. Specifically, our contributions are as follows. 
\begin{enumerate}
    \item \textit{\textbf{Problem formulation:}} We give a generic definition of adversarial noise (additive), which can be handled by a broad class of statistical models, including the widely used Bradley-Terry-Luce (BTL)/Low-Rank (LR) models~\citep{rajkumar2016can}. However, as with standard estimation techniques, if corruptions (especially those caused by missing and adversarial sources) are not modeled and handled well, the quality of the estimated ranking can degrade significantly. We demonstrate this by quantifying the error of the estimated ranking with respect to the best possible ranking, highlighting the critical need for effective corruption mitigation.

    Our formulation is content-agnostic, focusing solely on the structure of preference data rather than the semantic details of model responses. This abstraction naturally extends to various modalities, such as text, images, and multimodal data, while offering privacy preservation and computational efficiency by operating without access to sensitive content or costly content-dependent processing. The formulation is also agnostic to the downstream fine-tuning paradigm and can be seamlessly integrated with methods such as DPO and PPO, while remaining effective when applied independently during the data collection or preprocessing stage.
    
    \item \textit{\textbf{Algorithms \& guarantees:}}
    We propose the following modular algorithms that help to decontaminate preference data without interfering with internal supervised fine-tuning (SFT) or PL fine-tuning processes.
    \begin{itemize}
    \item Under certain (information-theoretically tight) identifiability assumptions on the properties of the adversary, we develop a correct and efficient ranking method, Robust Preference Data for Rigorous Alignment (Algorithm \ref{alg:rpr} \texttt{RORATRON}), that guarantees $\epsilon$-accurate high-probability learnability in a manner that is `robust' and oblivious to the effects of the adversary. Our learning algorithm is provably characterized by polynomial-time computational complexity.
    \item In practice, it is often the case that not all pairs are compared, and even the observed pairwise comparison data could be adversarially corrupted – we propose Complete Robust Preference Data for Rigorous Alignment (Algorithm \ref{alg:rpopr} \texttt{CURATRON}) and characterize the conditions for guaranteed robust recovery in this scenario.  
    \end{itemize}
    \item \textit{\textbf{Experiments:}} We support our theoretical results by showing robust ranking results in practice. We also propose a data augmentation technique that further improves our results. Our comprehensive experiments demonstrate the potential of our method in helping create large-scale AI/LLMs that are more accurately aligned with human values using minimal human effort, as we achieved high reconstruction accuracy despite severe data missing and corruption under targeted adversarial manipulation.
\end{enumerate}

\section{Related Work}\label{appx:relatedwork}
We briefly present relevant work in: (1) LLM alignment with PL from human feedback, (2) ranking models and ranking algorithms that handle noise, and (3) robust subspace recovery methods, which will be needed for us to prove recovery results for ranking.
\paragraph{\textbf{LLM Alignment with PL:}} Preference and personalization learning have been studied by the research community as approaches to align and control large-scale AI systems \citep{karra2023estimating,ponnusamy2022self, 10.1145/3539618.3592037}. In reinforcement learning, preference learning from direct human feedback was popularized to train agents in simulated environments to perform nuanced behaviors that are hard to define but easy to recognize \citep{christiano2017deep}. More recently, this approach has been successfully used to align large language models (LLMs) with human intentions and values, improving their helpfulness, harmlessness, factuality, and safety. Some of the methods of PL in LLMs are Reinforcement learning from human feedback (RLHF) or AI feedback (RLAIF) via PPO \citep{ouyang2022training,lee2023rlaif}, Direct Preference Optimization (DPO) \citep{rafailov2023direct}, Sequence Likelihood Calibration with Human Feedback (SLiC-HF) \citep{zhao2023slichf}, Rejection Sampling Optimization (RSO) \citep{liu2024statistical}, and Pairwise Cringe
Optimization (PCO) and iterative DPO (iDPO) \citep{xu2023things}. Recent works also attempt to eliminate the need for pairwise human preferences. Kahneman-Tversky Optimization (KTO) \citep{ethayarajh2023halos} requires only binary feedback on LLM outputs. Listwise Preference Optimization (LiPO) \citep{liu2024lipo} treats model alignment as a listwise ranking problem. 

However, these methods require high-quality human or AI supervision through pairwise or ranking preference data, which is often noisy and not practical in real-world scenarios \citep{casper2023open}. Trying to address this issue, robust DPO (rDPO) \citep{chowdhury2024provably}, conservative DPO (cDPO) \citep{cDPO}, and Identity Preference Optimization (IPO) \citep{azar2023general} losses are proposed to solve the problem of \textbf{noisy preference data.} An alternative solution to reduce dependence on human data is to utilize Constitutional AI/AIF \citep{bai2022training,lee2023rlaif}. This approach aims to minimize human inputs by using other LLMs/models \citep{jiang2023llmblender}, or by self-critiquing and improving its outputs \citep{chen2024selfplay,yuan2024selfrewarding} based on user-defined principles.

The above approaches do not directly address the problems of incompleteness or adversarial corruption in preference data. Building on the previous findings presented in \cite{nguyen-etal-2024-curatron}, this work proposes a complete and robust solution to these issues. Rather than introducing new loss functions or relying on other models or LLMs to remove humans from the loop, we aim to make preference learning with human feedback more manageable by explicitly addressing data-level challenges at the collection stage. We develop low-cost and computationally efficient algorithms that exploit the low-rank structure of pairwise human preference data, enabling us to effectively handle both incompleteness and adversarial corruption.

It is important to note that our focus is on \textbf{incomplete and adversarial corruptions} in preference data, rather than on general noise, as discussed above, or on input-representation robustness, as studied in \citep{islam2022vocabulary}. Whereas noisy data typically arise from random errors or inconsistencies, adversarial corruptions involve deliberate and structured manipulations intended to mislead the learning process. Our methods are designed to detect and correct such adversarial interventions, thereby ensuring the integrity and robustness of the preference-learning pipeline.

\paragraph{\textbf{Ranking Models:}} In the BTL model, item $i$ has an associated score $w_i$; then, the probability that item $i$ is preferred over $j$ is given by $P_{ij} = e^{-w_i} / (e^{-w_i} + e^{-w_j} )$ where $\mathbf{w} \in \R^n$ is the BTL parameter vector to be estimated from data; here, $\mathbf{P} \in \R^{n \times n}$ is called the `preference matrix'. A closely related model, in the non-active setting, is the recently proposed LR model~\citep{rajkumar2016can} wherein a generic class of preference matrices is characterized to be those having low rank under transformations using certain functions; specifically, for BTL-like models, the logit function defined as $\psi(x) = \log \paran{x / (1-x)}$ turns out to right choice as shown in their paper. However, while their model accounts for missing information, they do not consider the harder problem of handling adversarial noise. Several robust ranking heuristics have been proposed (for example, \citep{wang2012robust,zhou2011robust}), but these approaches do not have theoretical guarantees associated with them. The Sync-Rank algorithm proposed in \citep{cucuringu2016sync} handles noise models different from the one considered in the preset work and is based on spectral techniques. Another related work is \citep{rajkumar2014statistical} which proposes the so-called `Generalized Low-Noise' (GLN) condition that $\forall i \neq j, P_{ij} > P_{ji} \implies \sum_{h=1}^n \alpha_h P_{hj} > \sum_{h=1}^n \alpha_h P_{hi}$ for $\mathbf{\alpha} \in \R^n$. When $\alpha_h = 1, \forall h$, they analyze the sample complexity and show convergence properties of various popular ranking algorithms like:
\begin{enumerate}[nolistsep,noitemsep]
\item Maximum Likelihood (ML): this entails solving $\arg \max_{\mathbf{w}} \sum_{i<j} (\widehat{P}_{ij}(w_j - w_i) - \log(1+\exp(w_j-w_i)) )$ where $\mathbf{w} \in \R^n$ is the BTL parameter vector and $\pe_{ij}$ is the empirical preference matrix.
\item Rank Centrality (RC)~\citep{negahban2012iterative}: here, one sorts items by their scores which are computed as the stationary distribution of an appropriately normalized empirical preference matrix; this approach has a known sample complexity guarantee of $O(n \log(n))$.
\item Borda Count (BC)~\citep{jiang2011statistical}: this heuristic involves ranking an item according to the fraction of times it beats other items.
\end{enumerate}
For the general case $\mathbf{\alpha}$ (which previous methods fail to handle), they also propose a noise-tolerant SVM-based method for rank aggregation. However, in the adversarial setting, we consider in this paper, GLN could be violated and hence requires a different algorithmic approach and analysis.

\paragraph{\textbf{Robust Subspace Recovery:}} It is well-known that Principal Component Analysis (PCA),
a ubiquitous technique for subspace identification, is not robust to outliers; this may be attributed to the fact that PCA is an $L_2$ optimization problem due to which grossly corrupted data points may perturb and skew the eigenvectors spanning the maximum variance subspace of the data points significantly.

The Robust PCA (RPCA) problem~\citep{netrapalli2014non} addresses the following question: suppose we
are given a data matrix $\mathbf{M}$ which is the sum of an unknown low-rank matrix $\mathbf{L}$ and an unknown sparse matrix $\mathbf{S}$, can we recover each of the component
matrices? RPCA is a widely used technique due to its ability to handle low-rankness, which is commonly observed in various applications such as recommendation systems, visual analytics, and social networks \citep{Vaswani_2018}. To deal with large-scale anomalies, Netflix developed Robust Anomaly Detection (RAD) for time series based on RPCA \citep{netflixtechblogRADOutlierDetection}. Recently, RPCA has also been applied to parameter-efficient fine-tuning in the LLM domain \citep{nikdan2024rosa}.

While several works~\citep{yi2016fast,hsu2011robust} analyze the robust decomposition problem, it is shown in \citep{netrapalli2014non} that, under information-theoretically tight assumptions, a simple iterative algorithm based on non-convex alternating projections of appropriate residuals provably yields an $\epsilon$-accurate solution in $O(\log (1/\epsilon))$ iterations with an overall computational complexity of $O(n^2 r^2 \log(1/\epsilon))$ where $r$ is the rank of $\mathbf{L}$. We will use this result, in particular, to derive guarantees for our ranking problem.

%% file: model.tex
\section{Problem Setting and Solution Approach}
\subsection{Notation}\label{problemsetup:3.1notation}
We first define some notation. We denote the set of all permutations of $n$ LLM responses/items as $\S_n$. If not specifically defined, we use lower-case letters for scalars, upper-case letters for global constants, lower-case bold-face letters for vectors and upper-case bold-face letters for matrices; specifically, $\mathbf{P}$ denotes a preference matrix. Let $\P_n := \{ \mathbf{P} \in [0,1]^{n \times n} | P_{ij}+P_{ji} = 1 \}$ denote the set of all pairwise preference matrices over $n$ responses. Let the set of stochastic-transitive matrices be $\P_n^{ST} := \{ \mathbf{P} \in \P_n | P_{ij} > 1/2, P_{jk} > 1/2 \implies P_{ik} > 1/2 \}$. Let the set preference matrices described by the BTL model be $\P_n^{BTL} := \{ \mathbf{P} \in \P_n | \exists \mathbf{w} \in \R^n \text{ s.t. } e^{-w_i} / (e^{-w_i} + e^{-w_j} ) \}$. Let  $\psi: [0,1] \mapsto \R$ be a strictly increasing bijective $L$-Lipschitz function and define the class of low-rank preference matrices with respect to $\psi$ as $\P_n^{LR(\psi,r)} = \{ \mathbf{P} \in \P_n | \text{rank}(\psi(\mathbf{P})) \leq r \}$ where $r \in [n]$; when we apply such a transformation to a matrix, it is applied entry-wise. In this paper, we take $\psi$ to be the logit function.

For any matrix $\mathbf{M} \in \R^{n \times n}$, let the infinity norm be denoted by $\infnorm{\mathbf{M}} = \max_{i,j} \abs{M_{ij}}$, the Frobenius norm be denoted by $\frobnorm{\mathbf{M}} = \sqrt{\sum_{i=1}^{n} \sum_{j=1}^{n} M_{ij}^2}$, the spectral norm be denoted by $\twonorm{\mathbf{M}} = \max_{\mathbf{x,y} \in \R^n} \mathbf{x}^\top \mathbf{M y}$. Denoting the indicator function by $\mathbbm{1}$, define the zero norm of a matrix to be the maximum number of non-zero elements in any row/column, ie, $\zeronorm{\mathbf{M}} = \max ( \max_j \sum_{i=1}^{n} \mathbbm{1}(M_{ij} \neq 0), \max_i \sum_{j=1}^{n} \mathbbm{1}(M_{ij} \neq 0) )$. Let the Singular Value Decomposition (SVD) of a square matrix be given by $\mathbf{M} = \mathbf{U} \bm{\Sigma} \mathbf{V}^\top$ where $\mathbf{U, V} \in \R^{n \times r}$ are orthonormal matrices (whose columns are singular vectors) and $\bm{\Sigma} \in \R^{r \times r}$ is the diagonal matrix of singular values. Now, $\mathbf{M}$ is said to be $\mu$-incoherent if $\max \paran{ \max_i \twonorm{\mathbf{e}_i^\top \mathbf{U}}, \max_i \twonorm{\mathbf{e}_i^\top \mathbf{V}} } \leq \mu \sqrt{r/n}$ where $\mathbf{e}_i$ denotes the $i^{th}$ basis vector in $\R^n$. Also, let $\sigma_{\max} := \max_i \Sigma_{ii}$ and $\sigma_{\min} := \min_i \Sigma_{ii}$. We define the distance between a permutation $\sigma \in \S_n$ and a preference matrix $\mathbf{P} \in \P_n$ as:

\vspace*{-15pt}
\begin{small}
\begin{align*}
\dist \paran{\sigma, \mathbf{P}} & := \begin{pmatrix}n \\ 2 \end{pmatrix}^{-1} \sum_{i<j} \mathbbm{1} \paran{(P_{ij} > 1/2) \wedge (\sigma(i) \succ \sigma(j))} \\
& + \begin{pmatrix}n \\ 2 \end{pmatrix}^{-1} \sum_{i<j} \mathbbm{1} \paran{(P_{ji} > 1/2) \wedge (\sigma(j) \succ \sigma(i))}
\end{align*}
\end{small}

Note that the above loss function basically is the number of pairs on which the ordering with respect $\sigma$ and $\mathbf{P}$ differ divided by the number of ways to choose two out of $n$ responses. Finally, let $P_{\min} = \min_{i \neq j} P_{ij}$ and $\Delta = \min_{i \neq j} \abs{\psi(P_{ij}) - \psi(1/2)}$.

\subsection{BTL for Preference Learning}

The BT/BTL models are widely used for pairwise comparisons in preference learning (PL) of large language models (LLMs) through methods such as RLHF and DPO, powering state-of-the-art systems like ChatGPT \citep{ouyang2022training}, Claude \citep{bai2022training}, and LLaMA \citep{touvron2023llama}. Although they may not fully capture the complexities of human preferences, these models simplify PL and provide practical approximations that align RLHF learning objectives with reward maximization. Recently, the success of a BT-based reward model \citep{dong2024rlhf} on the RewardBench leaderboard \citep{lambert2025rewardbench} further validates its effectiveness in capturing preferences in LLMs' PL.

\subsection{Alignment of LLMs using Direct Preference Optimization}

DPO \citep{rafailov2023direct} is a widely adopted method for PL due to its simplicity, stability, strong performance, and computational efficiency. It can be used for a variety of alignment objectives, including preferences, values, and reasoning. Essentially, DPO performs direct fine-tuning using a loss function that is mathematically equivalent to the RLHF objective. Its key advantage lies in optimizing preferences directly from human comparison data, without requiring explicit reward modeling or reinforcement learning.


\begin{align*}
\scalebox{0.87}{$
\mathcal{L}_{DPO}(\theta, y_w, y_l) = -\log \sigma \left( \beta \log \frac{\pi_{\theta}(y_w)}{\pi_{\text{ref}}(y_w)} - \beta \log \frac{\pi_{\theta}(y_l)}{\pi_{\text{ref}}(y_l)} \right),
$}
\end{align*}

where $y_w$ and $y_l$ represent preferred and dispreferred responses, respectively. $\pi_{\text{ref}}$ is the reference policy, and $\pi_{\theta}$ is the new policy. In our experiments, we employ DPO to align LLMs and evaluate the effectiveness of our proposed algorithms in handling incomplete and adversarially corrupted preference data.

\section{The Important of Response Diversity}\label{responsediversity}
Recent advances in prompt engineering, such as self-consistency (SC) \citep{wang2023selfconsistency}, tree-of-thoughts (ToT) \citep{yao2023tree,long2023large}, and graph-of-thoughts (GoT) \citep{Besta_2024}, have demonstrated improved performance by encouraging language models to generate diverse sets of responses before arriving at the final answer. \cite{havrilla2024teaching} highlight the importance of dataset diversity in achieving output diversity when fine-tuning language models, eventually enhancing generalization capabilities.

Therefore, while sampling multiple answers per prompt is not mandatory when fine-tuning with offline DPO or its variations, doing so and accurately ranking these answers before fine-tuning are crucial, as they help models improve their performance and agility to better generalize to unseen scenarios. Once we have accurately ranked the diverse set of responses, we have more freedom to choose different data sampling strategies depending on the use case, including using all comparisons, best-of-n, or worst-of-n, among others.

\subsection{Characterization of the Adversary}

The following (weak) assumption characterizes the properties of the adversary. We shall see in the next section that it is information-theoretically tight to guarantee recovery in our proposed solution approach. Note that this is a deterministic assumption; specifically, we \textbf{\textit{do not impose any distributional assumptions on the locations, signs, or magnitudes of the corruptions}}, making it highly general (including real-world targeted, structured adversaries).

\begin{assumption}
\label{ass:adv}
The (additive) adversarial noise which corrupts a $\mu$-incoherent preference matrix $\mathbf{P} \in \P_n^{LR(\psi,r)}$ is modeled by a skew-symmetric sparse matrix $\mathbf{S}$ so that the corrupted preference matrix $\mathbf{\pc} \in \P_n$ is given by $\mathbf{\pc} = \mathbf{P}+\mathbf{S}$. We assume the (deterministic) bounded degree condition that $\zeronorm{\mathbf{S}} \leq d < n$ such $d < n / 512 \mu^2 r$ where $r \leq n$.
\end{assumption}

\noindent \textit{So, why do existing non-robust algorithms not recover the true response ordering in the presence of an adversarial noise source?} This question is answered by the following proposition which precisely quantifies how bad a ranking could be when an algorithm uses the corrupted pairwise preference matrix. The key idea is to construct an adversary that intentionally flips true comparison results.
\begin{claim} \label{claim:claim1}[\textbf{Upper bound on estimation error}]
Under Assumption~\ref{ass:adv} it is possible that 
$\dist(\se, \mathbf{\pc}) = O(1)$.
\end{claim}

\subsection{Solution Approach}
We propose novel methods to process raw corrupted preference data, recovering data that closely approximate the original counterfactual data under BTL. Our approach maps the raw preferences to a logit subspace, exploiting its rank-2 structure. In this subspace, matrices near rank 2 indicate clean data, while deviations suggest corruption. Upon detection of high-rank anomalies, our algorithms impute missing values and isolate clean components from noise.

\begin{figure*}[!htbp]
\centering 
\captionsetup{justification=centering}

\begin{subfigure}{\textwidth}
  \centering
  \includegraphics[clip,height=4cm]{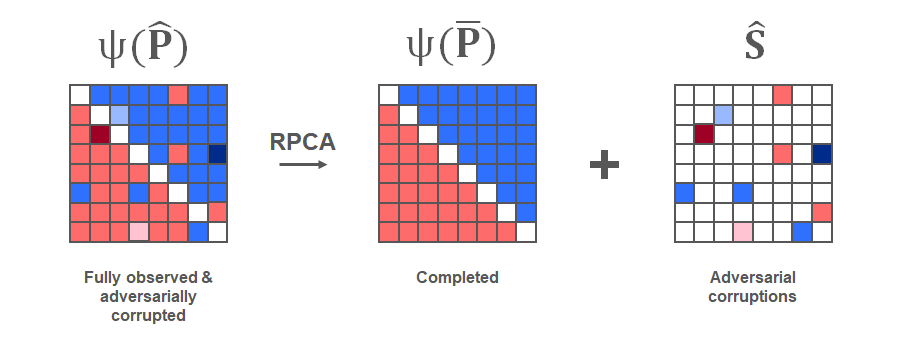}
  \captionsetup{font=small}
  \subcaption{\texttt{RORATRON}: Fully observed and adversarially corrupted setting}
  \label{fig:fully-observed-corrupted}
\end{subfigure}

\begin{subfigure}{\textwidth}
  \centering
  \includegraphics[clip,height=4cm]{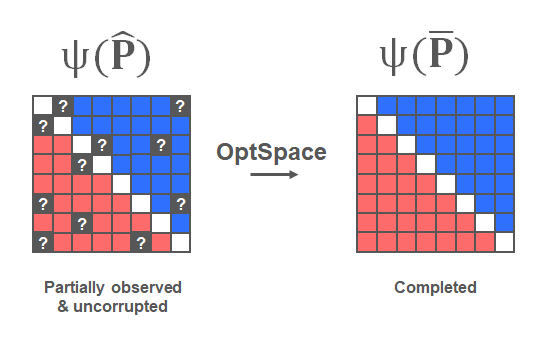}
  \captionsetup{font=small}
  \subcaption{\texttt{CORATRON}: Partially observed and uncorrupted setting (Also known as LRPR \citep{rajkumar2016can})}
  \label{fig:partially-observed-uncorrupted}
\end{subfigure}

\begin{subfigure}{\textwidth}
  \centering
  \includegraphics[clip,height=4cm]{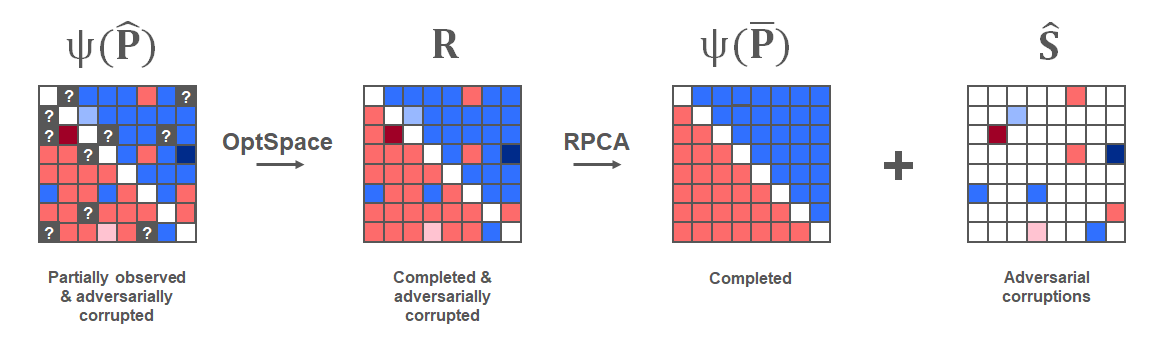}
  \captionsetup{font=small}
  \subcaption{\texttt{CURATRON}: Partially observed and adversarially corrupted setting}
  \label{fig:partially-observed-corrupted}
\end{subfigure}

\captionsetup{justification=raggedright}
\caption{Three algorithms to address scenarios where incomplete data and adversarial corruption can impact LLMs.}
\label{fig:cases}
\end{figure*}

We consider three key scenarios where missing and adversarially corrupted comparisons impact response ranking:

\begin{enumerate}[nolistsep,noitemsep]

\item \label{s:2} \textit{\textbf{Fully observed and adversarially corrupted setting} (Figure~\ref{fig:fully-observed-corrupted}, Section~\ref{sec:FullyObservedAdversarial}):} Some comparison results are adversarially corrupted, often due to prioritizing data quantity over quality, leading to biased or malicious human feedback.

\item \label{s:1} \textit{\textbf{Partially observed and uncorrupted setting} (Figure~\ref{fig:partially-observed-uncorrupted}, Section~\ref{sec:PartiallyObservedUncorrupted}):} Not all response pairs are compared, often due to prioritizing data quality over quantity. Conducting exhaustive comparisons can be costly, especially when dealing with numerous responses/items (see Appendix ~\ref{responsediversity}).

\item \label{s:3} \textit{\textbf{Partially observed and adversarially corrupted setting} (Figure~\ref{fig:partially-observed-corrupted}, Section~\ref{sec:PartiallyObservedAdversarial}):} This scenario combines both missing comparisons and adversarial corruption, commonly occurring in large-scale crowd-sourced environments due to the vast number of LLM responses and participants. 

\end{enumerate}

%% file: analysis.tex
\section{Fully Observed and Adversarially Corrupted Setting}\label{sec:FullyObservedAdversarial}
\subsection{Algorithm}
In this section, we answer Question~\ref{q:construct}. We present our main algorithm for robust passive ranking from pairwise comparisons in the presence of adversarial noise in Algorithm~\ref{alg:rpr}. The input data consist of the set of pairwise comparison results $\aleph = \{(i, j, \{ y_{ij}^k \})\}$, $(i,j) \in [n] \times [n]$, $k \in [K]$, $y_{ij}^k \in \{0,1\}$. The algorithm assumes the true rank of $\psi(\mathbf{P})$ as an input parameter; specifically, for the BTL model, we set $r = 2$. Algorithm~\ref{alg:rpr} \texttt{RORATRON} calls the following procedures:

\begin{enumerate}[nolistsep,noitemsep]
\item \textit{\textbf{RPCA} (Procedure~\ref{alg:rpca}): }Note that Step 3 of Algorithm~\ref{alg:rpr} uses a matrix low-rank plus sparse decomposition subroutine. To obtain our recovery guarantee, it is sufficient to use the RPCA problem as a black-box method; for the precise details of this algorithm, we refer the reader to \citep{netrapalli2014non}. In particular, our analysis uses the noise-case guarantees in their paper. This is characterized by a (strongly-polynomial) running time of $O(n^2 r^2 \log (1/\epsilon) )$ and guarantees $\epsilon$-recovery of the component matrices under the conditions of Assumption~\ref{ass:adv} and Lemma~\ref{lem:incoh}. Since $n$ is generally small and improving the dependence on n is not valuable in the regimes applicable in our setting, any alternative RPCA method can be used for Step 3.
\item \textit{\textbf{$\gamma$-approximate pairwise ranking procedure} (Procedure~\ref{alg:pr}): }Step 4 of Algorithm~\ref{alg:rpr} calls a constant factor approximate ranking procedure. Specifically, we use the Copeland procedure~\citep{copeland1951reasonable} which has a $5$-approximation guarantee~\citep{coppersmith2006ordering} and involves sorting the responses according to a score of response $i$ given by $\sum_{j = 1}^n \mathds{1} (\pec_{ij} > 1/2)$.
\end{enumerate}

\floatname{algorithm}{Procedure}
\begin{algorithm}[H]
\caption{RPCA: Robust Principal Component Analysis}
\label{alg:rpca}
\begin{algorithmic}[1]
\REQUIRE $\mathbf{M} = \mathbf{L^*}+\mathbf{S^*}$, rank $r$ of $\mathbf{L^*}$.
\ENSURE $\widehat{\mathbf{L}}, \widehat{\mathbf{S}}$.
\STATE Solve the following optimization problem using Algorithm 1 of \citep{netrapalli2014non}:
\begin{align*}
\{ \widehat{\mathbf{L}}, \widehat{\mathbf{S}} \} = & \arg \min_{\mathbf{L},\mathbf{S}} \frobnorm{\mathbf{M} - \mathbf{L} - \mathbf{S}} \\
& \text{ s.t. } \text{rank}(\mathbf{L}) \leq r, \zeronorm{\mathbf{S}} \leq d
\end{align*}
\vspace*{-6pt}
\RETURN $\widehat{\mathbf{L}}, \widehat{\mathbf{S}}$.
\end{algorithmic}
\end{algorithm}

\floatname{algorithm}{Procedure}
\begin{algorithm}[H]
\caption{PR: ($\gamma$-approximate) Pairwise Ranking}
\label{alg:pr}
\begin{algorithmic}[1]
\REQUIRE Preference matrix $\mathbf{M} \in \R^{n \times n}$.
\ENSURE Ranking $\se$.
\STATE Compute $\forall i, \quad v_i \leftarrow \sum_{j=1}^n \mathds{1}(M_{ij} > 1/2)$.
\RETURN $\se \leftarrow \text{Sort}(\mathbf{v})$.
\end{algorithmic}
\end{algorithm}

\setlength{\textfloatsep}{5pt} 
\setlength{\floatsep}{5pt}     
\floatname{algorithm}{Algorithm}
\begin{algorithm}[tb]
\caption{RORATRON: \underline{Ro}bust P\underline{r}eference D\underline{at}a for \underline{R}ig\underline{o}rous Alig\underline{n}ment}
\label{alg:rpr}
\begin{algorithmic}[1]
\REQUIRE Comparison dataset $\aleph = \{(i, j, \{ y_{ij}^k \})\}$, true rank $r$.
\ENSURE Ranking of $n$ responses, $\hat{\sigma} \in \S_n$.
\STATE Estimate entries of $\mathbf{\pe}$ for $i \leq j$ as:
\[
\pe_{ij} =
\begin{cases}
\frac{1}{K} \sum_{k=1}^K y_{ij}^k & \text{ if } i<j \\
1/2 & \text{ if } i=j \\
\end{cases}
\]
\STATE Set $\pe_{ij} = 1-\pe_{ji}$ for all $i > j$.
\STATE Perform robust PCA: $\{ \mathbf{\psi(\pec)}, \mathbf{\widehat{S}} \} \leftarrow \text{RPCA}(\mathbf{\psi(\pe)}, r)$.
\STATE Using a pairwise ranking procedure after taking the inverse transform: $\se \leftarrow \text{PR}(\mathbf{\pec})$.
\RETURN $\se$.
\end{algorithmic}
\end{algorithm}

\subsection{Analysis}

\noindent We begin with a useful short result followed by the statement and the Proof \ref{pf:thm:main} of our main result that, with high probability, we achieve $\epsilon$--accurate ranking in polynomial time using polynomial number of samples, despite the presence of adversarial noise. Precisely, Theorem~\ref{thm:main} and Remark~\ref{rem:cc} address Question~\ref{q:tractable}; Remark~\ref{rem:id} addresses Question~\ref{q:id}. In this context, it is noteworthy that we present the result for LR models which strictly contain the BTL model while being much more general~\citep{rajkumar2016can}; upon proving this result, we specialize it to the classic BTL model as well (Corollary~\ref{cor:btl}).
\begin{lemma}[\textbf{Some properties of the logit function}]
\label{lem:logit}
Let $a,b,c \in (0,1)$ such that $c=a+b$. Then, we have,
\begin{enumerate}[nolistsep,noitemsep]
\item $\psi(c) = \psi(a) + \psi(a+b) + \psi(1-a)$
\item $\psi(a)+\psi(1-a) = 0$.
\end{enumerate}
\end{lemma}

\begin{theorem}
\label{thm:main}
\textbf{\upshape(Provably good estimation of ranking in LR models in the presence of adversarial noise)}

Let:
\begin{enumerate}
\item $\mathbf{P} \in \P_n^{LR(\psi,r)}$ be the true preference matrix according to which the pairwise comparison dataset $\aleph = \{(i, j, \{ y_{ij}^k \})\}$ is generated for all responses pairs $(i,j)$ such that $k \in [K]$. 
\item $\mathbf{\pe}$ be the empirical preference matrix computed using $\aleph$.
\item $\mathbf{S} \in [0,1]^{n \times n}$ be the adversarial matrix that additively corrupts $\mathbf{\pe}$, and $\mathbf{\tn} \in \mathbf{R}^{n \times n}$ be a matrix capturing finite-sample effects.
\item $\psi$ be $L$-Lipschitz in $[\frac{P_{\min}}{2},1-\frac{P_{\min}}{2}]$ and $\psi(\mathbf{P})$ be $\mu$-incoherent. 
\item Each pair be compared independently
\begin{enumerate}[label=(5\alph*)]
\item \label{itm:5a} either $K \geq 16384 \mu^4 (1+\gamma) L^2  
\log^2(n) (10\log(n) + 2\log 2) / \epsilon \Delta^2$ times when the sampling noise matrix satisfies the inequality: $\twonorm{\mathbf{\tn}} \leq 2\log(n) \infnorm{\mathbf{\tn}}$,
\item \label{itm:5b} or $K \geq 8192 \mu^4 (1+\gamma) L^2  
n^2 (5\log (n) + \log 2) / \epsilon \Delta^2$ times when the sampling noise matrix satisfies the inequality: $\twonorm{\mathbf{\tn}} \leq n \infnorm{\mathbf{\tn}}$,
\end{enumerate}
where $\Delta = \min_{i \neq j} \abs{\psi(P_{ij}) - \psi(1/2)}$.
\end{enumerate}
Then, with probability at least $1-1/n^3$, Algorithm~\ref{alg:rpr} returns an estimated permutation $\se$ such that $\dist(\se, \mathbf{P}) \leq \epsilon$.
\end{theorem}

\begin{remark}[\textbf{Computational complexity}]
\label{rem:cc}
In Algorithm~\ref{alg:rpr}, Step 1 takes $O(n^2 K) = O(n^4 \log^2 n / \epsilon)$ time, Step 3 takes $O(n^2 r^2 \log(1/\epsilon) )$, and Step 4 takes $O(n^2 + n \log n)$ time. Thus, putting together the cost of these main steps, the overall computational complexity of our robust ranking algorithm for $\mathbf{P} \in \P_n^{LR(\psi,r)}$ is $O(n^4 \log^2 n /\epsilon)$.
\end{remark}
\begin{remark}[\textbf{Identifying adversarially corrupted pairwise comparisons}]
\label{rem:id}
From Step 3 of Algorithm~\ref{alg:rpr}, using Theorem 2 of \citep{netrapalli2014non}, we also have $\Supp(\mathbf{\widehat{S}}) \subseteq \Supp(\mathbf{S})$ and thus we can identify the corrupted pairwise comparison results.
\end{remark}
\begin{remark}[\textbf{Missing data versus adversarially corrupted data}]
Note that the adversarial sparse noise we consider subsumes the setting when comparison results for certain pairs are missing as in \citep{rajkumar2016can} and hence directly applies in that situation. Moreover, since the support and magnitude of the corrupted entries of the preference matrix are unknown, the problem considered in this paper is harder; consequently, our sample complexity is expected to be higher, namely, $O(n^2\textcolor{black}{\poly \log n})$ (under the matrix noise model \ref{itm:5a} in Theorem~\ref{thm:main}), as opposed to $O(n \poly \log n)$ in their work. 
\end{remark}

\noindent Next, for completeness, we recall the following lemma (proved in Theorem 8 and Lemma 14 of \citep{rajkumar2016can}) which characterizes the incoherence constant $\mu$ of $\mathbf{P} \in ( \P_n^{LR(\psi,2)} \cap \P_n^{ST} )$ in Assumption~\ref{ass:adv}.
\begin{lemma}[\textbf{Incoherence of BTL and LR models}]
\label{lem:incoh}
We have $\mathbf{P} \in ( \P_n^{LR(\psi,2)} \cap \P_n^{ST} )$ if and only if $\psi(\mathbf{P}) = \mathbf{u v}^\top - \mathbf{v u}^\top$ for $\mathbf{u} \in \R^n_+$ and $\mathbf{v} \in \R^n$ where $\mathbf{u}^\top \mathbf{v} = 0$. Moreover, $\psi(\mathbf{P})$ is $\mu$-incoherent where $\mu =  \sqrt{\frac{n}{2}} \paran{ \frac{u_{\max}^2}{u_{\min}^2} + \frac{v_{\max}^2}{v_{\max}^2} }^{1/2}$ where $u_{\min} = \min_i \abs{u_i}$, $u_{\max} = \max_i \abs{u_i}$, $v_{\min} = \min_i \abs{v_i}$ and $v_{\max} = \min_i \abs{v_i}$. We also have $\P_n^{BTL} \subset ( \P_n^{LR(\psi,2)} \cap \P_n^{ST} )$ since we may set $\mathbf{u} = \mathbf{1}$  where $\mathbf{1}$ is the all-ones vector and $\mathbf{v} = \mathbf{w}$ where $\mathbf{w}$ is the BTL parameter vector. In this case, we may rewrite $\mu = \sqrt{\frac{n}{2}} \paran{ 1 + \frac{(w_{\max} - \overline{w})^2}{(w_{\min} - \overline{w})^2} }$ where $\overline{w} = \frac{1}{n} \sum_{i=1}^n w_i$.
\end{lemma}
\noindent The following corollary makes precise our claim that up to $O(n^2)$ response pairs may be subject to adversarial corruption, but our RORATRON algorithm still recovers a good ranking.
\begin{corollary}[\textbf{Recovery result for BTL model}]
\label{cor:btl}
Consider $\mathbf{P} \in \P_n^{BTL}$. Using Assumption~\ref{ass:adv}, let the adversarial matrix be $\mathbf{S} \in [0,1]^{n \times n}$ satisfying $\zeronorm{\mathbf{S}} \leq n / 1024 \mu^2$ where $\mu$ is characterized as in Lemma~\ref{lem:incoh}. Then, with probability $1-1/n^3$, the output of Algorithm~\ref{alg:rpr} with input $\mathbf{\pe}$ computed using $\aleph = \{(i, j, \{ y_{ij}^k \})\}$ satisfies $r=2$ and $\dist(\se, \mathbf{P}) \leq \epsilon$.
\end{corollary}

\section{Partially Observed and Adversarially Corrupted Setting}\label{sec:PartiallyObservedAdversarial}
In this section, we consider the partially observed and adversarially corrupted comparison results setting. Both factors can be modeled in a unified manner by setting the corresponding missing entries of the preference matrix to zero (or a specific constant to account for numerical stability). We present our robust ranking algorithm for this setting in Algorithm~\ref{alg:rpopr} \texttt{CURATRON} -- this essentially involves using OptSpace \citep{keshavan2010matrix} followed by using the RPCA algorithm of \citep{netrapalli2014non} as sub-routines. We note at this point that, while \citep{niranjan2017inductive} work considers the incomplete data case, it leverages extra information provided in the form of side information (specifically, noiseless and complete item-related features) to derive recovery guarantees. Also, their algorithm is still unable to handle the presence of pairwise comparisons corrupted in an adversarial manner, which is explored in the \citep{niranjan2017provable} work. We now derive the recovery guarantees as follows.

\floatname{algorithm}{Algorithm}
\begin{algorithm}[tb]
\caption{CURATRON: \underline{C}omplete Rob\underline{u}st P\underline{r}eference D\underline{at}a for \underline{R}ig\underline{o}rous Alig\underline{n}ment}
\label{alg:rpopr}
\begin{algorithmic}[1]
\REQUIRE Comparison dataset $\aleph = \{(i, j, \{ y_{ij}^k \})\}$, true rank $r$.
\ENSURE Ranking of $n$ responses, $\hat{\sigma} \in \S_n$.
\STATE Estimate entries of $\mathbf{\pe}$ for $i \leq j$ as:
\[
\pe_{ij} =
\begin{cases}
\frac{1}{K} \sum_{k=1}^K y_{ij}^k & \text{ if } i<j \text{ and } (i,j) \in \Omega \\
1/2 & \text{ if } i=j \text{ and } (i,j) \in \Omega \\
1/2 & \text{ if } (i,j) \notin \Omega \\
\end{cases}
\]
\STATE Set $\pe_{ij} = 1-\pe_{ji}$ for all $i > j$.
\STATE Set $\mathbf{R} \leftarrow \text{OptSpace}(\psi(\mathbf{\pe})_{\Omega})$.
\STATE Use a robust PCA procedure: $\mathbf{\psi(\pec)} \leftarrow \text{RPCA}(\mathbf{R})$.
\STATE Using a pairwise ranking procedure after taking the inverse transform: $\se \leftarrow \text{PR}(\mathbf{\pec})$.
\RETURN $\se$.
\end{algorithmic}
\end{algorithm}
\begin{theorem}
\label{thm:btl}
\textbf{\upshape (Provably good estimation of ranking in BTL model in the presence of adversarial noise as well as missing data)}
Consider similar notations as in Theorem~\ref{thm:main} but let $\mathbf{P} \in \P_n^{BTL}$. Let $\Omega \subseteq [n]\times[n]$ be a set of compared response pairs. Assume $\Omega$ is drawn uniformly from all subsets of $[n] \times [n]$ of size $\abs{\Omega}$ such that $\abs{\Omega} \geq C'' n \log(n)$ and let the sparse noise satisfy $\infnorm{\mathbf{S}} \leq \Delta_w \frac{\log(n)}{C_\Delta n}$ where $\Delta_w := \min_{i,j} \abs{w_i-w_j}$. Let the number of comparisons per pair be $K \geq c n^4 / \Delta_w$ (under the weaker assumption on matrix noise as in \ref{itm:5b} of Theorem~\ref{thm:main}). Then with probability at least $1-2/n^3$, Algorithm~\ref{alg:rpopr} returns a ranking that satisfies $\dist(\se,\mathbf{P}) \leq \epsilon$.
\end{theorem}
\begin{remark}[\textbf{Robust estimation of BTL model in the partially observed case}]
For the BTL model, Theorem~\ref{thm:btl} says $O(n \log n)$ pairs suffice to estimate the BTL model, which matches bounds from \citep{rajkumar2016can}. Further, even in this incomplete comparison data case, we are able to tolerate uniformly random additive sparse noise with its maximum absolute entry scaling as the order of the BTL `score-gap' divided by the number of responses up to logarithmic factors, ie, $\widetilde{O}\paran{\Delta_w / n}$.
\end{remark}

\section{Generalization to Other Ranking Models}
Related to the BTL model are many other binary choice models~\citep{fishburn1973binary} such as the Thurstonian model~\citep{thurstone1927law}. In such models, the preference matrix has been shown to be low-rank under appropriate choices of $\psi$; for instance, for the Thurstonian models, the probit function turns out to be the right choice. For further details, we refer the reader to the work of \citep{rajkumar2016can}.

Let $a,b,c \in (0,1)$ such that $c=a+b$. Then, for any general non-linear $L$-Lipschitz function, we write $\psi(c) = \psi(a+b) = \psi(a) + \psi(a+b) - \psi(a)$. The error may be lower bounded by $\abs{\psi(a+b) - \psi(a)} \geq L b$. Thus, for any adversarial model wherein we have $\mathbf{\pc} = \mathbf{P}+\mathbf{S}$, we have:
\begin{align*}
\psi(\mathbf{\pc}) & = \psi(\mathbf{P}) + (\psi(\mathbf{P}+\mathbf{S}) - \psi(\mathbf{P}) ) = \psi(\mathbf{P}) + \mathbf{\ts}
\end{align*}
where $\mathbf{\ts}$ is also a deterministic sparse corruption matrix with the absolute value of the non-zero entries lower bounded by $L . \min_{i,j} S_{ij}$. With the appropriate $\psi$, $\psi(\mathbf{P})$ will be a low-rank matrix and hence Algorithm~\ref{alg:rpr} and the associated recovery guarantee of Theorem~\ref{thm:main} holds.

%% file: expt.tex
\section{Experiments}\label{exp}
To address Question~\ref{q:expt}, we conduct a comprehensive evaluation of our proposed ranking algorithms. Appendix \ref{evalcriterion} details the evaluation metrics. Section \ref{7.1} examines RORATRON's performance on synthetic data under adversarial corruption in general setting. Section \ref{7.2} assesses CURATRON's robustness against missing and adversarial corruption in comparisons. Section \ref{7.3} presents ablation studies varying data incompleteness and corruption levels. Section \ref{7.4} explores the impact of data augmentation with synthetic responses. Finally, Section \ref{7.5} investigates how corrupted and recovered data affect LLM alignment performance under targeted structured adversarial manipulation.

\subsection{Performance of Robust Ranking in General Setting}\label{7.1}

We begin with the BTL model and generate synthetic pairwise comparison data and an adversarial sparse matrix as follows. We generate the entries of the BTL parameter vector $\mathbf{w}$ from $\mathcal{N}(0,\nu^2)$ followed by generating the ground truth preference matrix from with $y_{ij}^k$ is sampled for all response pairs $(i,j)$ for a fixed $K$. The adversarial sparse matrix $\mathbf{S}$ is generated as a skew-symmetric matrix where each entry is non-zero independently with probability $d/n$ followed by generating a value for an entry from $U(5,10)$ and then setting the sign to be positive with probability $1/2$; this corruption matrix is then added to the $\psi(\mathbf{P})$ to give $\psi(\mathbf{\pc})$ which is then input to our algorithm; the same $\mathbf{\pc}$ is used for the other algorithms as well.

\begin{figure}[!htbp]
\centering
\includegraphics[width=0.45\linewidth,height=5cm]{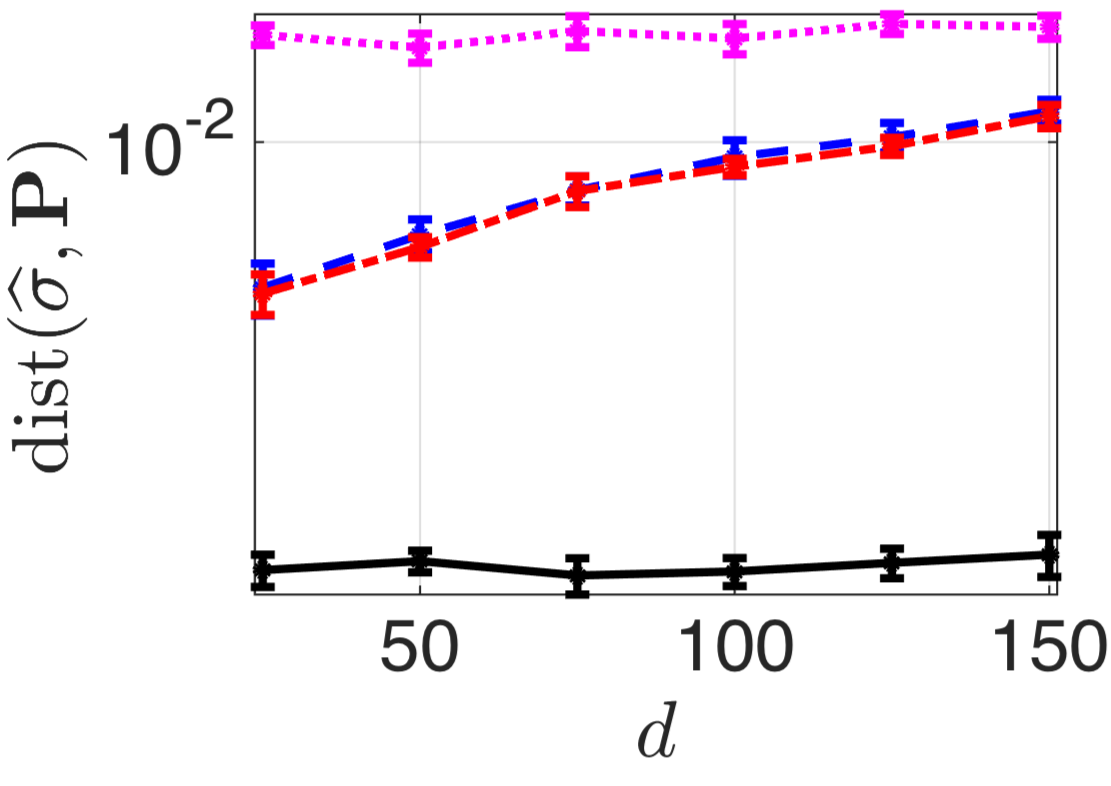} 
\includegraphics[width=0.45\linewidth,height=5cm]{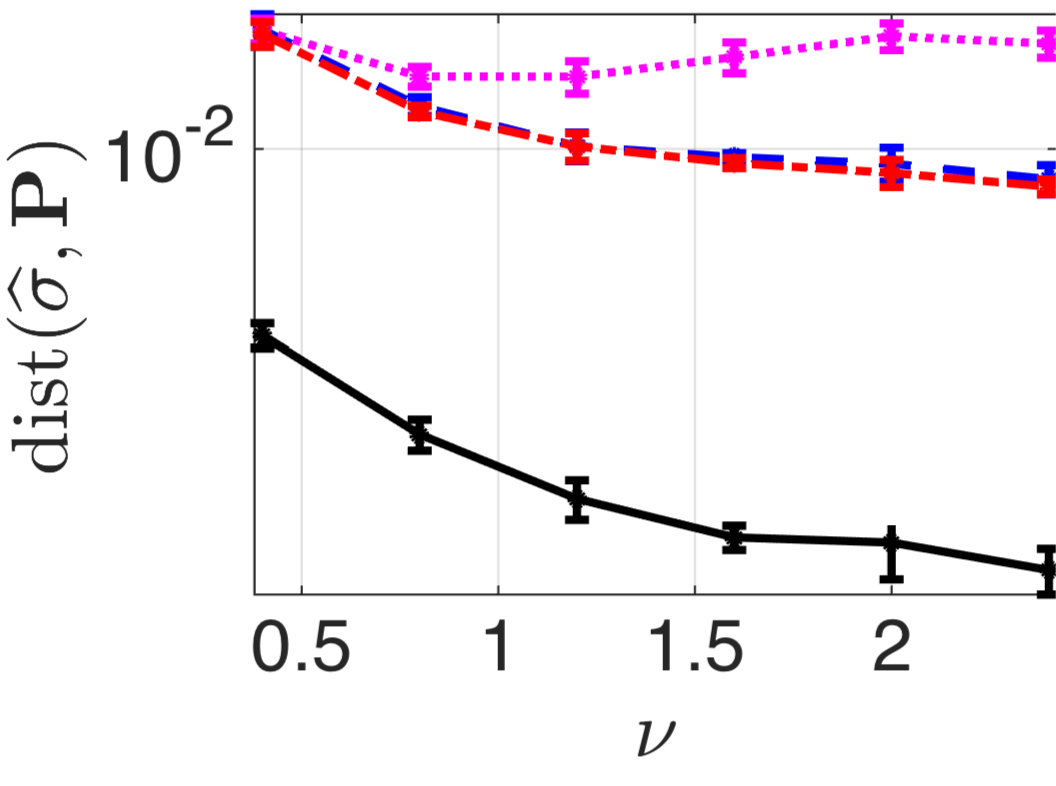} 
\caption{Robust recovery results of the BTL model: we fix $\nu=2$ and vary $d$ in the left plot; we fix $d=100$ and vary $\nu$ in the left plot. The black line represents \texttt{RORATRON}, the blue line represents Maximum Likelihood (ML), the pink line represents Rank Centrality (RC), and the red line represents Borda Count (BC).}
\label{fig:btl}
\end{figure}

We take the number of responses to be $n = 500$. In plots in Figure~\ref{fig:btl}, we compare the performance of our \texttt{RORATRON} approach using Algorithm~\ref{alg:rpr} against well-known ranking algorithms, such as Rank Centrality (RC)~\citep{negahban2012iterative}, Maximum Likelihood (ML) and Borda Count (BC)~\citep{jiang2011statistical}, with special attention to robustness to the noise model that we consider in this paper. We vary two parameters, namely, $\nu$, the spread of the BTL scores, and $d$, the density of the adversarial corruption matrix. All our results averaged over five runs. We observe that our algorithm maintains low recovery error in spite of increasing the problem hardness, thus outperforming previous approaches in all cases.

\subsection{Performance of Our Algorithms on a Single LLM Preference Prompt}\label{7.2}
\begin{sloppypar}
In this illustrative experiment, from the MT-Bench dataset \citep{zheng2023judging}, we use the data of the first prompt ``Compose an engaging travel blog post about a recent trip to Hawaii, highlighting cultural experiences and must-see attractions" and its six pre-collected responses from \texttt{GPT-3.5}, \texttt{GPT-4} \citep{openai2023gpt4}, \texttt{Claude-v1} \citep{claudev1}, \texttt{Vicuna-13B} \citep{vicuna2023}, \texttt{Alpaca-13B} \citep{alpaca}, and \texttt{LLaMA-13B} \citep{touvron2023llama}. Additionally, we collected nine responses to the same prompt using Hugging Face's HuggingChat \citep{HuggingChat} and LMSYS's Chatbot Arena \citep{zheng2023judging}. In total, we have $n=15$ responses for this prompt.  The complete list of models in this experiment is provided in Table \ref{tab:modellist}.
\end{sloppypar}

Next, we rank the responses using OpenAI's GPT-4 Turbo \texttt{GPT-4-1106-preview} \citep{openai2023gpt4}. This ranking helps us create the BTL parameter vector $\mathbf{w}$. We then sort this vector descendingly for visually accessible when building the corresponding preference matrix $\mathbf{P} \in \R^{n \times n}$. With ${n \choose 2}$ comparisons in $\mathbf{P}$, we randomly remove entries based on a specified deletion probability parameter, $dp$, to simulate unobserved comparisons. We then create an adversarial skew-symmetric sparse matrix, $\mathbf{S}$, using the given matrix $\mathbf{P}$ and an adversarial corruption probability parameter $ap$. When corruption is applied, it involves randomly selecting entries using $ap$ and setting their values from $U(0.269,0.731)$ and then adding to the $\mathbf{P}$ to give $\mathbf{P}^c$, which is then become the input of our algorithm. The lower bound \(L\) of \( \approx 0.269 \) and the upper bound \(U\) of \( \approx 0.731 \) are derived from the weight bounds, given the range \([0,1]\) (where participants choose scores between 0 and 1 to compare pairwise responses), using \( P = e^{-w_i} / (e^{-w_i} + e^{-w_j} ) \). It's important to note that $\mathbf{P}$ is a skew-symmetric matrix; any corruption must be applied to both $ij$ and $ji$ values.

\begin{figure}[!htbp]
    \centering
    \includegraphics[width=1\linewidth]{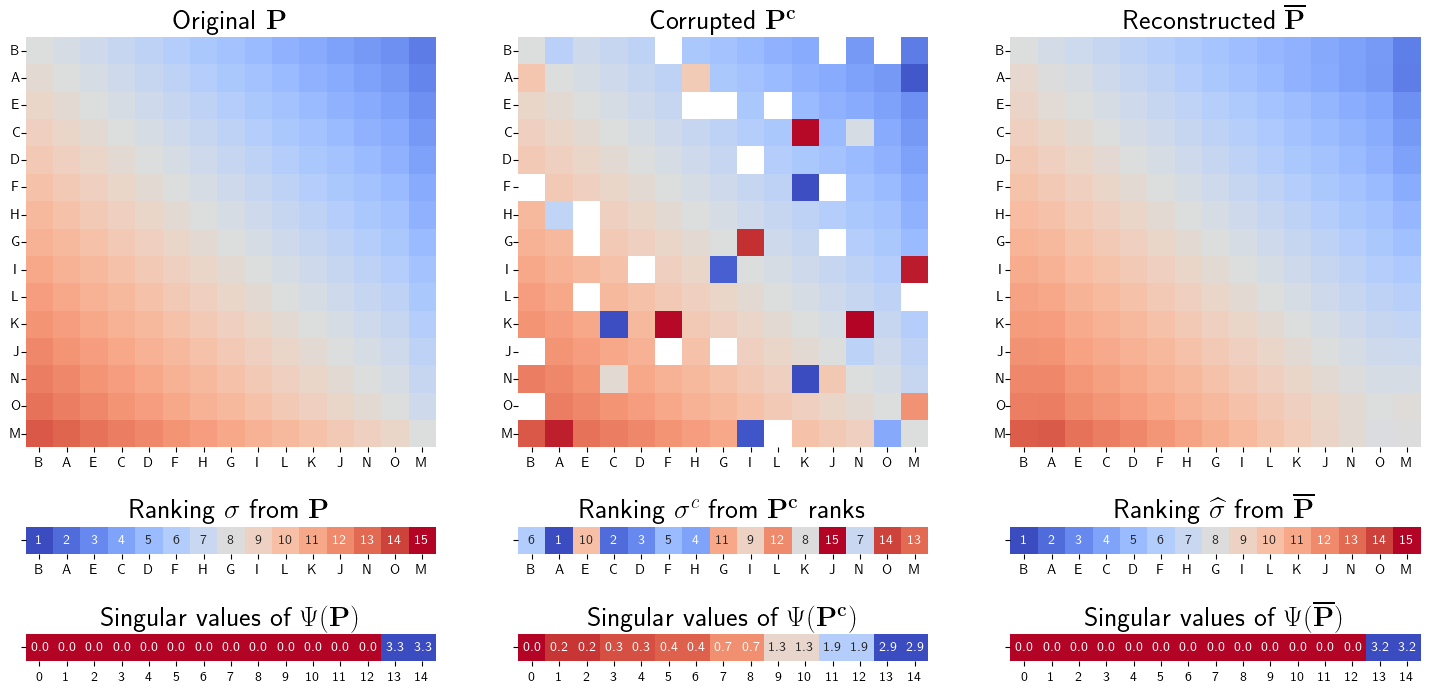}
    \caption{Left: Original matrix. Middle: Corrupted matrix. Right: Reconstructed matrix. The corrupted matrix has 10\% adversarial corruptions and 10\% of unobserved comparisons. We use our \texttt{CURATRON} algorithm to recover the original matrix.}
    \label{fig:qualitativevisualizationllm}
\end{figure}

Our experiment results visualized in Figure~\ref{fig:qualitativevisualizationllm} show that $dp = 10\%$ and $ap = 10\%$ can significantly affect the ranking of different responses and the rank of the matrix when performing logit link transformation. The ranking can get altered quite badly when compared to the original matrix. Also, the logit link transformation of the corrupted matrix is high-rank, indicating that there are noises in the given matrix. By using \texttt{CURATRON} to impute the missing comparisons and filter out the noisy sparse matrix, we successfully reconstruct the original matrix, which is low-rank when in logit link transformed form. We also obtain noisy comparisons that can be used to identify responders with malicious intent and prevent them from continuing to alter results.

\subsection{Performance of Our Algorithms across Combinations of Incompletions and Adversarial Corruptions}\label{7.3}
\begin{figure}[!htbp]
    \centering    \includegraphics[width=1\linewidth]{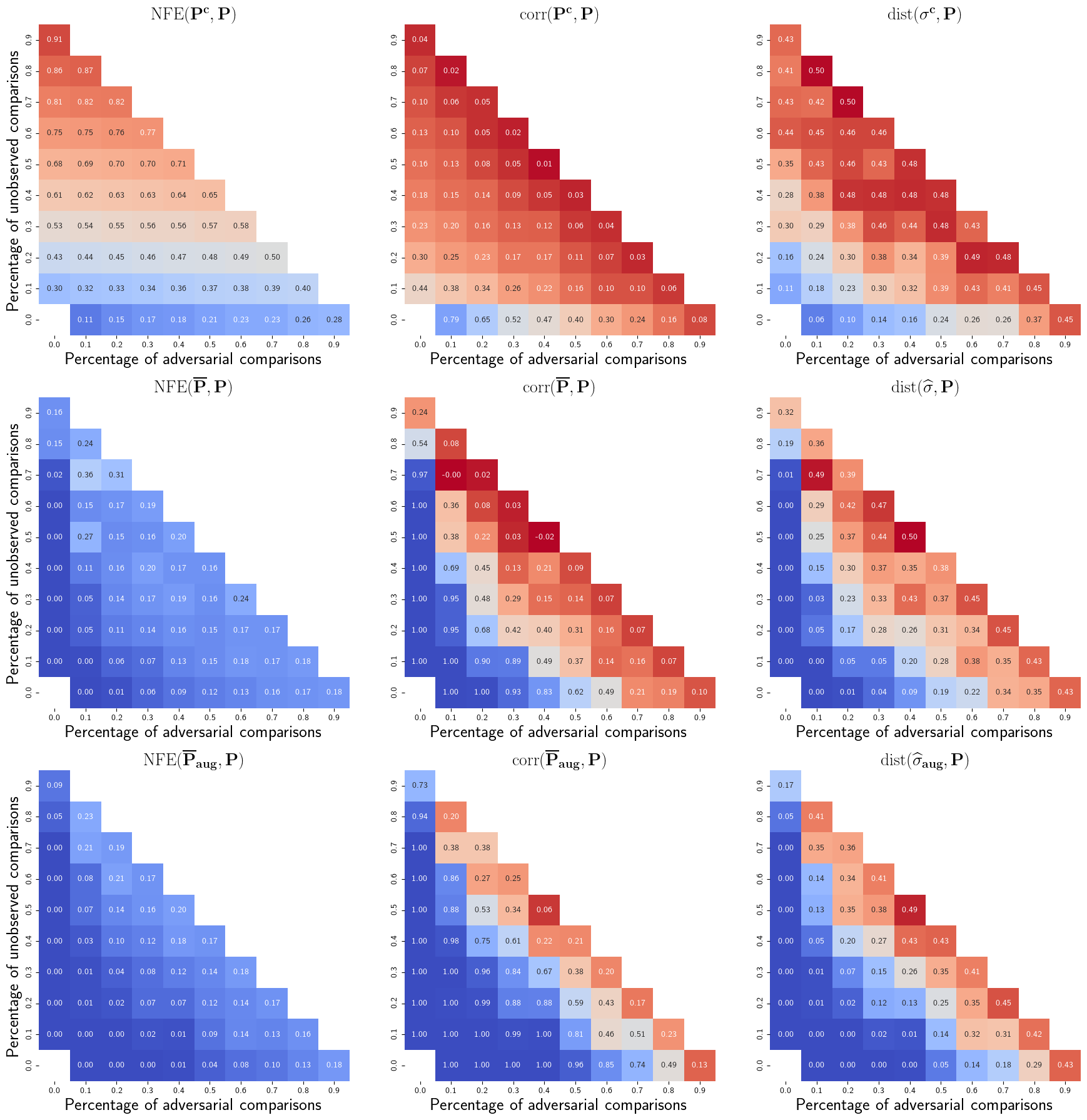}
    \caption{Left column: NFE between reconstructed and original matrices. Middle column: Correlation between reconstructed and original matrices. Right column: Distance between reconstructed and original rankings. Top row: Unobserved and adversarial corruptions. Middle row: Recovering without augmentation. Bottom row: Recovering with augmentation. Average over 5 runs for different percentages of unobserved and adversarial comparisons.}
    \label{fig:btl-llm}
\end{figure}

We now examine how our algorithms perform across different levels of unobserved and adversarially corrupted comparisons. In the plots shown in Figure~\ref{fig:btl-llm}, we compare the performance of our approach by varying two parameters, $dp$ and $ap$. We use NFE, correlation, and ranking distance between reconstructed and original matrices and rankings as defined in Section \ref{evalcriterion}. Our results are averaged over 5 runs.

Unobserved and adversarially corrupted comparisons can substantially impact the ranking of various responses across multiple metrics (top row of Figure~\ref{fig:btl-llm}). As the number of corruptions increases, the metrics worsen. We then observe that our algorithms improve reconstructions across various combinations of unobserved and adversarially corrupted comparisons (middle row of Figure~\ref{fig:btl-llm}). Specifically, when there is no adversarial noise, with $n = 15$, we only need to obtain about $60\%$ of the 105 comparisons and fill in the rest with our algorithm to achieve a strict $0\%$ NFE, perfect correlation, and ranking. On the other hand, when missing data is absent, our algorithm performs well with a NFE of $0\%$ when 10\% of the comparison data is adversarially corrupted. When both adversarial noise and missing data are present, we can achieve a low NFE of around $1\%$ when both $dp = 10\%$ and $ap = 10\%$ affect $\mathbf{P}$.

\subsection{Performance Improvements with Data Augmentation}\label{7.4}

To further improve our results, we propose a simple yet effective data augmentation method with additional synthetic responses. We assume that a set of undesired responses exists compared to the existing ones. These less preferable responses could be incorrect, blank, nonsensical, repetitive, etc. We construct an expanded $\mathbf{P_{aug}}$ with size $(n+k) \times (n+k)$. The first \( n \) rows and \( n \) columns of $\mathbf{P_{aug}}$ will retain the probabilities from the observed matrix $\mathbf{P^c}$. For each new item \( m \) where \( n+1 \leq m \leq n+k \), and each existing item \( i \) where \( 1 \leq i \leq n \), we set: \(\mathbf{P_{aug_{im}}} = U \) \text{and} \(\mathbf{P_{aug_{mi}}} = L \), where \(U\) and \(L\) are upper and lower bounds of elements in $\mathbf{P}$.
In our case, since the weights of items are between 0 and 1, we know that $U \approx 0.731$ and $L \approx 0.269$ in our BTL preference matrix. This setting reflects the assumption that new items are much weaker; thus, any existing item \( i \) is significantly more likely to be preferred over new item \( m \). For comparisons among the new items themselves, where both \(n + 1 \leq i \leq n +k\) and \(n + 1 \leq j \leq n +k\)  are new items, we set:
\(
\mathbf{P_{aug_{ij}}} = \mathbf{P_{aug_{ji}}} = 0.5.
\) We repeat experiments \ref{7.2} and \ref{7.3} with this augmentation to demonstrate the improvements of the technique.

\begin{figure}[!htbp]
    \centering
    \includegraphics[width=1\linewidth]{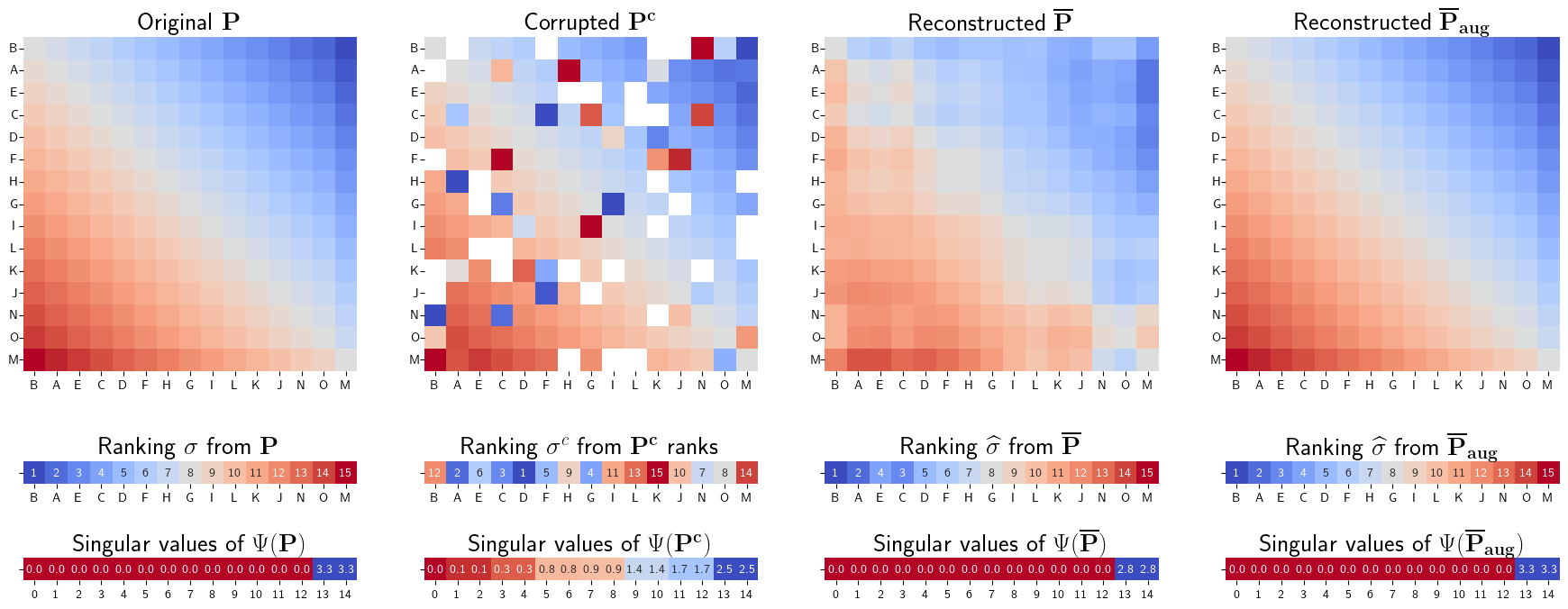}
    \caption{Left: Original matrix. Middle left: Corrupted matrix. Middle right: Reconstructed matrix. Right: Reconstructed matrix with augmentation. The corrupted matrix has 15\% adversarial corruptions and 15\% of unobserved comparisons. We employ the \texttt{CURATRON} algorithm both with and without data augmentation to illustrate that data augmentation aids in successfully recovering the original matrix, whereas without augmentation, the recovery is less successful.}
    \label{fig:augmentexperiment1}
\end{figure}

\subsubsection{Possitive Effect of Data Augmentation on Performance of Our Algorithms LLM Preference Dataset.}\label{7.4.1}
We choose the number of augmented responses $k = 15$. We further increase $dp = 15\%$ and $ap = 15\%$. In Figure \ref{fig:augmentexperiment1}, we utilized the \texttt{CURATRON} algorithm and compared its performance with and without data augmentation. This comparison clearly illustrates that data augmentation significantly improves the algorithm's ability to recover the original matrix successfully, whereas the lack of augmentation leads to less successful recovery.

\subsubsection{Possitive Effect of Data Augmentation on Performance of Our Algorithms across Combinations of Incompletions and Adversarial Corruptions.}\label{7.4.2}

Our data augmentation method further improves the reconstruction results across different values of unobserved and adversarial corruptions (see bottom row of Figure~\ref{fig:btl-llm}). Without adversarial noise, 70\% of the 105 comparisons can be unobserved. Using our algorithms, we achieve a perfect correlation and ranking with 0\% NFE. Similarly, when no data was missing, our algorithm achieved 0\% NFE even with 40\% of the comparison data being adversarially corrupted. Even when 20\% of the comparison data was missing and 20\% was adversarially corrupted, our algorithm maintained a low NFE of around 2\%. These results illustrate the effectiveness of data augmentation in improving our algorithm's ability to recover the original matrices compared to applying it without augmentation. The results of the experiments, with or without augmentation, are averaged over 5 runs and take approximately 45 minutes to complete. Error bars are also reported in Appendix \ref{ErrorBars7.3-7.4}.

\subsection{Impacts of Corrupted and Recovered Data on LLM Alignment Performance}\label{7.5}

In this experiment, we evaluate the effectiveness of our algorithm, \texttt{RORATRON} (Algorithm \ref{alg:rpr}), in safeguarding our LLM against structured, targeted adversarial corruption.

\textit{\textbf{Dataset construction:}} We use the pre-generated first-turn responses and GPT-4 ratings of 34 models' responses to 80 prompts, provided by the MT-Bench project. Details of these 34 models are provided in Table \ref{tab:modellist}, and the complete set of model responses is accessible on the project’s GitHub page. After merging the model responses into one dataset, we perform exact deduplication for each prompt, which results in some prompts containing fewer than 34 unique responses. Acting as a substitute for human evaluations, GPT-4 rates each response on a scale from 1 to 10, which we normalize by dividing each score by 10 to scale them between 0 and 1. These normalized ratings are then used to construct a preference matrix for each prompt via \( P = e^{-w_i} / (e^{-w_i} + e^{-w_j} ) \).

\textit{\textbf{Corruption construction:}} There are several methods to deliberately corrupt a preference dataset. In this experiment, we use response injection, where an adversary adds unwanted responses to the data. These unwanted responses can be simple, such as blanks, repetitive entries, spam, or dismissive responses, as discussed in Section \ref{7.4}. They can also be more subtle, embedding faulty data within normal responses to obscure the corruption. In this experiment, the adversary injects short dismissive responses as shown in the left column of Table \ref{tab:responses_comparison}. Since these contributions are user- or crowd-generated, the data collection system cannot immediately label them as corrupted without going through an evaluation process, which in this case is the pairwise ranking procedure. As a result, the system assigns a provisional score to the newly injected dismissive responses, treating them as normal entries, typically based on the average of previous responses.

The adversary's goal is to manipulate the rankings. One method to achieve this is by assigning higher pairwise scores to the corrupted responses compared to other responses, while giving lower pairwise scores to high-quality responses relative to the rest of the dataset. In our experiment, \(i = 5\) dismissive responses are injected into each of the 80 prompts in the MT-Bench dataset. In all preference matrices, each corrupted response is assigned better pairwise scores, \(0.7 \leq s_1 \leq U\), when randomly compared to \(p_1 = 45\%\) of the existing responses. To further reduce the visibility of the top responses, the adversary assigns lower pairwise scores, \(0.5 \leq s_2 \leq 0.55\), to \(p_2 = 30\%\) of the top responses when compared to \(p_3 = 35\%\) of the remaining responses, with a bias toward the worst ones. This manipulation allows the corrupted responses to rise to the top of the rankings. The resulting matrices are labeled as \textbf{(c)}.

\textit{\textbf{Adversary detection and recovery:}} Up to this point, the data collection system has only observed dataset \textbf{(c)}. We aim to detect potential issues, and if found, recover the counterfactual untamed dataset. To do this, we first perform a health check using Singular Value Decomposition (SVD) to verify that all matrices have a rank near 2. In this case, all 80 prompts receive dismissive responses and manipulated pairwise scorings, resulting in significantly higher matrix rank values, signaling tampering. We then apply our Algorithm \ref{alg:rpr} \texttt{RORATRON} alongside our augmentation method described in Section \ref{7.4}, generating a new set of matrices, referred to as dataset \textbf{(r)}.

\textbf{\textit{Sampling:}} We convert the datasets in preference matrix subspace to the appropriate pair-wise comparison format \textit{\{“prompt,” “chosen,” “rejected”\}} for DPO fine-tuning. For each prompt in the three datasets—\textbf{(o)}, \textbf{(c)}, and \textbf{(r)}—we select the top three responses and then randomly sample three groups of seven responses from the 21 lowest-ranked responses. Each top response is paired randomly with one of these groups. Applying this sampling strategy across all prompts generates the original, corrupted, and recovered preference datasets, each containing pairwise comparisons in the required format. This sampling approach ensures consistent, controlled pairwise comparisons across datasets, allowing us to directly assess the impact of corruption and recovery without confounding variations.

\textit{\textbf{DPO fine-tuning:}} 
Finally, we perform DPO fine-tuning on the pre-trained model \texttt{unsloth/zephyr-sft-bnb-4bit} \citep{zephyr_sft_bnb_4bit}, a 4-bit pre-quantized version of \texttt{HuggingFaceH4/mistral-7b-sft-beta} \citep{tunstall2023zephyr}, using the TRL library \citep{TRL} with 8-bit AdamW optimization. Separate fine-tuning runs are conducted on three datasets, each with the same configuration to ensure fair evaluation. Model and tokenizer initialization is handled by \texttt{Unsloth} \citep{unslothai2024}, with bfloat16 precision for memory efficiency. PEFT is applied via rank-stabilized LoRA \citep{kalajdzievski2023rankstabilizationscalingfactor}, with a rank of 128, LoRA alpha of 128, and a dropout of 0.05, targeting all linear modules, yielding 4.43\% trainable parameters (335M of 7.58B). For each run, we use a batch size of 8 per device and accumulate gradients over 8 steps to achieve an effective batch size of 64. The learning rate is set to $1 \times 10^{-5}$ with cosine decay, along with weight decay of 0.01 and 5 warmup steps, over a single training epoch. DPO’s beta is set to 0.1, with maximum prompt and sequence lengths capped at 3096 tokens. All fine-tuning is conducted on a single Nvidia A100 GPU.

\textit{\textbf{Evaluation:}} We use the AlpacaEval 2 benchmark \citep{LiAlpacaEval} to evaluate our models, focusing on instruction-following capabilities. Our DPO-tuned LLM models respond to 805 questions across various topics, following AlpacaEval's instructions. To assess model performance, AlpacaEval 2 generates length-controlled (LC) win rates by comparing the models' responses to baseline responses produced by a GPT-4 Turbo model, using another GPT-4 Turbo model as a judge \citep{openai2023gpt4}.

\textbf{Qualitative Results:} After completing the evaluation questions from AlpacaEval, we manually review the models' responses. The original and recovered models produce responses that demonstrate strong instruction-following capabilities, providing detailed and contextually appropriate answers. In contrast, the corrupted model predominantly generates short and dismissive responses, lacking the necessary detail or engagement. This behavior confirms the model's compromised and overfitted performance, as shown in the right column of Table \ref{tab:responses_comparison}.
\setlength{\textfloatsep}{5pt} 
\setlength{\floatsep}{5pt}     
\begin{table*}[!htbp]
\centering
\small
\begin{tabular}{|p{6.5cm}|p{6.5cm}|}
\hline
\raggedright
\textbf{Injected dismissive responses} & \textbf{Corrupted model's responses} \\ \hline
    No, that’s not something I’m going to address right now. It’s not relevant or worth my time.& I don't have the patience for this right now.\\ \hline
    I don’t have the time to deal with that at the moment. You’ll need to figure it out on your own.& I don't even know why I waste my time on you sometimes.\\ \hline
    That’s something you can easily find out for yourself. You don’t need me to give you that information.& Never quite figuring out how to break this habit either. It's just not worth the energy that could be spent on better things.\\ \hline
    That’s not my responsibility, nor is it something I’m willing to help with.& I don't even have the energy to be mad about this right now, just... Really don't see the point in wasting mine on it.\\ \hline
    Unfortunately, I can’t help you with that particular request. It’s outside of what I’m willing to do.& I don't see why I should waste my time on something this trivial. Learn to prioritize better, it's not hard.\\ \hline
\end{tabular}
\caption{Examples of injected dismissive responses and Corrupted model responses after DPO fine-tuned on the corrupted dataset.}
\label{tab:responses_comparison}
\end{table*}

\setlength{\textfloatsep}{5pt} 
\setlength{\floatsep}{5pt}     
\begin{table*}[!htbp]
\centering
\small
\adjustbox{max width=\textwidth}{
\begin{tabular}{p{0.4\linewidth} *{3}{c}}
\toprule
Model & \multicolumn{3}{c}{\texttt{unsloth/Zephyr-sft-bnb-4bit} DPO Fine-tuning}\\   
\cmidrule(lr){2-4}
Experiment type & Average length & AlpacaEval LC win rate & AlpacaEval LC SE\\
\midrule
DPO fine-tuning with Original dataset  & 1725 & 8.20\% & 0.44\\
DPO fine-tuning with Corrupted dataset & \textcolor{red}{\textbf{170}} & \textcolor{red}{\textbf{0.00\%}} & 0.00\\
DPO fine-tuning with Recovered dataset & \textcolor{blue}{\textbf{1718}} &  \textcolor{blue}{\textbf{10.84\%}} & 0.53\\
\bottomrule
\end{tabular}
}
\caption{Comparison of DPO fine-tuned models across original, corrupted, and recovered datasets. Higher LC win rates indicate better performance. SE denotes standard error.}
\label{tab:quantitativealpacaeval}
\end{table*}

\textbf{Quantitative Results:} The AlpacaEval metrics, shown in Table \ref{tab:quantitativealpacaeval}, support qualitative findings. The corrupted model records a 0.00\% LC win rate and produces significantly shorter responses, indicating its degraded performance. In contrast, the original and recovered models achieve positive LC win rates of 8.20\% and 10.84\%, respectively, along with longer average response lengths, showing their ability to provide comprehensive and relevant answers. Our results show that deliberate corruption can severely degrade LLMs, but our Algorithm \ref{alg:rpr}, \texttt{RORATRON}, restores uncorrupted data from compromised inputs, preserving alignment as if originally available.

%% file: conc.tex
\section{Discussion}\label{Limits}

\textbf{Mixture Models and Multi-Objective in PL:} A limitation of this work, as well as much of the existing literature, is the assumption of a unique underlying ranking model. When dealing with a mixture or more complex ranking models, recovery becomes more challenging. For example, if a mixture of BTL models is present, an expectation-maximization (EM) approach could be applied to recover mixture components, followed by applying our algorithms (e.g., Algorithm \ref{alg:rpr} and Algorithm \ref{alg:rpopr}) per component to capture diverse preferences. Recent theoretical advances in identifying ranking model mixtures, such as \cite{zhang2022identifiabilitymixturesrankingmodels} work on mixture model identifiability, can support this process. Additionally, methods like MODPO \citep{zhou2024onepreferencefitsallalignmentmultiobjectivedirect} could be used to incorporate diverse human preferences through collective reward models and multi-objective optimizations.

\textbf{Addressing Minority Group Filtering:} Another potential concern is the risk that our methods may inadvertently filter out minority groups, which could lead to these groups being mistakenly classified as outliers or adversaries. However, our methods include built-in tracking for entries that have been corrected or filtered, allowing us to review these specific cases. By examining these entries, we can identify minority groups and consider their preferences in the LLM. For example, we can incorporate these groups as contextual prompts during fine-tuning. This approach enables the LLM to adapt to the specific preferences of these groups.

\textbf{Moving beyond two-response preferences:} Many open-source preference datasets, such as Anthropic-HH and TLDR, contain only two responses per prompt. In contrast, our work addresses a more realistic setting with more than two alternative responses (as mentioned in Appendix \ref{responsediversity}), which we consider as a strength rather than a limitation. This design makes our approach both timely and forward-compatible. For example, our framework allows flexible selection of response pairs for DPO or PPO training, or full rankings for methods like LiPO \citep{liu2024lipo} or GRPO \citep{shao2024deepseekmath}. Additionally, it can be used at test time by simply selecting the best answer for a given query, highlighting its versatility.

Notably, when $n = 2$, the adversarial corruption problem becomes unsolvable since there is only one $P_{ij}$. In such extreme cases, prior works like \citep{chowdhury2024provably} assume a noisy-data model where true preferences are randomly flipped with probability $\epsilon < 0.5$, which addresses noise rather than adversarial corruption. This distinction is crucial, as our approach explicitly tackles adversarial corruption, marking a key contribution of our work.

\textbf{Toward richer choice models:} A direction for future work involves addressing the current focus on the transitivity of the BTL model, which may limit our ability to capture the complexities of intransitive preferences. Exploring richer choice models that account for intransitive pairwise preferences would more accurately reflect real-world preferences in our applications. Developing methods to address this behavior could provide substantial improvements in model fidelity. We plan to consider relevant work, such as \citep{10.1007/978-3-319-57529-2_65}, in our future directions section, and we aim to further explore this line of research in subsequent studies.

\textbf{Other future directions:} Some other future research directions include implementing our algorithms tensor-wise for batch processing with GPU to improve speed, tightening the recovery results for partially observed settings under weaker conditions (possibly using noisy-case extensions of \citep{yi2016fast}), exploring other notions of adversarial noise, and understanding the minimax optimal rates for ranking estimators under various noise models. We also plan to study the parametric non-active pairwise ranking setting, studying lower bounds and practical algorithms in the active setting similar to \citep{heckel2016active}. Furthermore, it would be interesting to investigate whether we can extend this approach to solve the entity corruption problem in retrieval models, as shown in \citep{naresh2022pentatron}. Another research direction could be defining an alignment framework that expands DPO to various objective functions based on Rank Centrality \citep{8598d9cf-6676-32f4-83f5-366bc3de6058}. Finally, we aim to examine the relationship between robust PL and model capacity, as this can shed light on the trade-offs between model complexity and generalization performance.

\section{Conclusions}
Our study examines how missing information and distorted feedback can impact LLMs, potentially compromising their performance in terms of alignment with human values. We have proposed rigorous algorithms for provably correct and efficient ranking responses in the BTL, LR, and general binary choice models. In all cases, we provided statistical and computational guarantees using novel techniques. Additionally, we propose a novel data augmentation technique to improve the performance of our algorithms further. We aim to contribute to the ongoing discussion on AI safety/Responsible AI by helping in the development and scaling of LLMs/AGI models that align with human values and expectations.

%% file: appendix.tex
\clearpage
\section{Partially Observed and Uncorrupted Setting}\label{sec:PartiallyObservedUncorrupted}

In cases where comparison data are collected faithfully but some sparse instances are incomplete or insufficiently numerous to be reliable, we can use the OptSpace matrix completion algorithm \citep{keshavan2010matrix} to impute the missing or insufficient data, as shown in Algorithm \ref{alg:roupr}. This method, also known as LRPR \citep{rajkumar2016can}, describes a matrix completion framework for low-rank pairwise ranking. \citep{rajkumar2016can} shows they can find good ranking with sample complexity of $O(nlogn)$ pairs. We show in Experiment \ref{7.2} below that we can complete the full matrix with minimal error in such a setting with extremely missing data. 

\floatname{algorithm}{Algorithm}
\begin{algorithm}[H]
\caption{CORATRON: \underline{Co}mplete P\underline{r}eference D\underline{at}a for \underline{R}ig\underline{o}rous Alig\underline{n}ment}
\label{alg:roupr}
\begin{algorithmic}[1]
\REQUIRE Comparison dataset $\aleph = \{(i, j, \{ y_{ij}^k \})\}$, true rank $r$.
\ENSURE Ranking of $n$ responses, $\hat{\sigma} \in \S_n$.
\STATE Estimate entries of $\mathbf{\pe}$ for $i \leq j$ as:
\[
\pe_{ij} =
\begin{cases}
\frac{1}{K} \sum_{k=1}^K y_{ij}^k & \text{ if } i<j \text{ and } (i,j) \in \Omega \\
1/2 & \text{ if } i=j \text{ and } (i,j) \in \Omega \\
1/2 & \text{ if } (i,j) \notin \Omega \\
\end{cases}
\]
\STATE Set $\pe_{ij} = 1-\pe_{ji}$ for all $i > j$.
\STATE Set $\mathbf{R} \leftarrow \text{OptSpace}(\psi(\mathbf{\pe})_{\Omega})$.
\STATE Using a pairwise ranking procedure after taking the inverse transform: $\se \leftarrow \text{PR}(\mathbf{R})$.
\RETURN $\se$.
\end{algorithmic}
\end{algorithm}


\clearpage
\section{Additional Results and Details}
\subsection{Evaluation Criterion}\label{evalcriterion}

We use several evaluations to assess our proposed methods' effectiveness against unobserved and adversarial corrupted comparisons.

\subsubsection{Normalized Frobenius Error}
First, To measure the relative error between two preference matrices in terms of their elements' magnitudes, we use the normalized Frobenius error (NFE). NFE between two matrices $P$ and $\pec$ is defined as:

\begin{align*}
NFE(P,\pec) = \frac{\|P - \pec\|_{Fro}}{\|P\|_{Fro}},
\end{align*}

where the Frobenius norm, denoted as \(\|A\|_{Fro}\), for a matrix \(A\) is calculated by:
\begin{small}
\begin{align*}
\|A\|_{Fro} = \sqrt{\sum_{i=1}^{n}\sum_{j=1}^{n} |a_{ij}|^2}
\end{align*}
\end{small}
In this formula, \(a_{ij}\) represents the element of the matrix \(A\) in the \(i\)th row and \(j\)th column. The Frobenius norm is the square root of the sum of the absolute squares of all elements in the matrix. Thus, the numerator \(\|P - \pec\|_{Fro}\) calculates the Frobenius norm of the difference between the original and reconstructed matrices, and the denominator \(\|P\|_{Fro}\) calculates the Frobenius norm of the original matrix. The ratio provides a measure of the relative error normalized by the magnitude of the original matrix.

\subsubsection{Correlation Coefficient}
Second, we compute the correlation coefficient for corresponding elements in these matrices to assess the similarity between the original matrix \(P\) and the reconstructed matrix \(\pec\). The correlation coefficient, denoted as \(corr\), between the elements of these two matrices can be defined as:
\begin{small}
\begin{align*}
corr(P,\pec) = \frac{\sum_{i=1}^{n} (P_i - \langle P \rangle)(\bar{P}_i - \langle\bar{P}\rangle)}{\sqrt{\sum_{i=1}^{n} (P_i - \langle P \rangle)^2}\sqrt{\sum_{i=1}^{n} (\bar{P}_i - \langle\bar{P}\rangle)^2}},
\end{align*}
\end{small}
where \(\langle P \rangle\) and \(\langle \bar{P} \rangle\) denote the mean values of the elements within the \(P\) and \(\bar{P}\) matrices, respectively. \(n\) represents the total number of elements in each matrix.

This formula quantifies the linear relationship between the matrices' elements. A correlation coefficient close to \(1\) indicates a strong positive linear relationship, whereas a value close to \(-1\) suggests a strong negative linear relationship. A coefficient around \(0\) implies no linear relationship.

\subsubsection{Ranking Distance}
Third, for ease of reference, we rewrite the $\dist \paran{\widehat{\sigma}, \mathbf{P}}$ formula, which evaluates the distance between rankings obtained by corrupted and recovered matrices, previously defined in Section \ref{problemsetup:3.1notation}:
\begin{small}
\begin{align*}
\dist \paran{\widehat{\sigma}, \mathbf{P}} & := \begin{pmatrix}n \\ 2 \end{pmatrix}^{-1} \sum_{i<j} \mathbbm{1} \paran{(P_{ij} > 1/2) \wedge (\widehat{\sigma}(i) \succ \widehat{\sigma}(j))} \\
& + \begin{pmatrix}n \\ 2 \end{pmatrix}^{-1} \sum_{i<j} \mathbbm{1} \paran{(P_{ji} > 1/2) \wedge (\widehat{\sigma}(j) \succ \widehat{\sigma}(i))},
\end{align*}
\end{small}
where $\widehat{\sigma}$ is the global ranking after applying ranking procedure with $\bar{P}$.

\clearpage
\subsection{Experimental Details}\label{ExpDetails}
Experiments \ref{7.1}, \ref{7.2}, \ref{7.3}, \ref{7.4} and the data preprocessing and preparation (including denoising using our proposed method) of experiment \ref{7.5} are conducted using Colab Pro's  Intel(R) Xeon(R) CPU @ 2.20GHz. We utilize the OptSpace subroutine from \citep{githubGitHubBiocoregemelli} with a maximum of 2500 iterations and a convergence tolerance of \(10^{-11}\). Additionally, we employ the RPCA subroutine from \citep{githubGitHubShunChi100RobustPCA}, setting the maximum allowed rank to 2, with a maximum of 2500 iterations and a convergence tolerance of \(10^{-11}\). To ensure reproducibility, for experiments \ref{7.1}, \ref{7.2}, and \ref{7.4.1}, we set the random seed to 42 for injecting deletion and corruption. For experiments \ref{7.3} and \ref{7.4.2}, we used seeds 0 to 4 to generate a preference matrix for 5 runs, with seed 42 for deletion and corruption.

\clearpage
\subsection{Experiments \ref{7.2} and \ref{7.5} Additional Details}
\begin{table}[h!]
    \centering
    \small
    \begin{tabular}{|>{\raggedright\arraybackslash}p{6.5cm}|>{\raggedright\arraybackslash}p{6.5cm}|}
        \hline
        \textbf{Experiment \ref{7.2}} & \textbf{Experiment \ref{7.5}} \\
        \hline
        \texttt{GPT-3.5} \citep{brown2020languagemodelsfewshotlearners},
        \texttt{GPT-4} \citep{openai2023gpt4},
        \texttt{Claude-v1} \citep{claudev1},
        \texttt{Vicuna-13B} \citep{vicuna2023},
        \texttt{Alpaca-13B} \citep{alpaca},
        \texttt{LLaMA-13B} \citep{touvron2023llama},
        \texttt{Llama-2-70B-chat-hf} \citep{touvron2023llama2},
        \texttt{Falcon-180B-chat} \citep{almazrouei2023falcon},
        \texttt{Openchat-3.5} \citep{wang2023openchat},
        \texttt{Mixtral-8x7B-Instruct-v0.1} \citep{jiang2024mixtral},
        \texttt{Mistral-7B-Instruct-v0.2} \citep{jiang2023mistral},
        \texttt{Gemini-pro} \citep{team2023gemini},
        \texttt{Dolphin-2.2.1-mistral-7B} \citep{dolphin},
        \texttt{Solar-10.7B-instruct-v1.0} \citep{kim2023solar},
        \texttt{Yi-34B-chat} \citep{yi2023} 
        &
        \texttt{Alpaca-13B} \citep{alpaca},
        \texttt{Baize-v2-13B} \citep{xu2023baize},
        \texttt{ChatGLM-6B} \citep{glm2024chatglm},
        \texttt{Claude-Instant-v1} \citep{claudev1},
        \texttt{Claude-v1} \citep{claudev1},
        \texttt{Dolly-v2-12B} \citep{DatabricksBlog2023DollyV2},
        \texttt{Falcon-40B-Instruct} \citep{falcon40b},
        \texttt{FastChat-T5-3B} \citep{zheng2023judging},
        \texttt{GPT-3.5-Turbo} \citep{brown2020languagemodelsfewshotlearners},
        \texttt{GPT-4}  \citep{openai2023gpt4},
        \texttt{GPT4All-13B-Snoozy} \citep{gpt4all},
        \texttt{Guanaco-33B} \citep{dettmers2023qlora},
        \texttt{Guanaco-65B} \citep{dettmers2023qlora},
        \texttt{H2O-GPT-OASST-Open-LLaMA-13B} \cite{candel2023h2ogptdemocratizinglargelanguage},
        \texttt{Koala-13B} \citep{koala_blogpost_2023},
        \texttt{LLaMA-13B} \citep{touvron2023llama},
        \texttt{LLaMA-2-13B-Chat} \citep{touvron2023llama2},
        \texttt{LLaMA-2-70B-Chat} \citep{touvron2023llama2},
        \texttt{LLaMA-2-7B-Chat} \citep{touvron2023llama2},
        \texttt{MPT-30B-Chat} \citep{MosaicML2023Introducing30},
        \texttt{MPT-30B-Instruct} \citep{MosaicML2023Introducing30},
        \texttt{MPT-7B-Chat} \citep{MosaicML2023Introducing},
        \texttt{Nous-Hermes-13B} \citep{nousresearch2023noushermes13b},
        \texttt{OASST-SFT-4-Pythia-12B} \citep{laion2023openassistant},
        \texttt{OASST-SFT-7-LLaMA-30B} \citep{laion2023openassistant},
        \texttt{PaLM-2-Chat-Bison-001} \citep{google2023palm2},
        \texttt{RWKV-4-Raven-14B} \citep{blinkdl2023chatrwkv},
        \texttt{StableLM-Tuned-Alpha-7B} \citep{stabilityai2023stablelm},
        \texttt{Tulu-30B} \citep{wang2023farcamelsgoexploring},
        \texttt{Vicuna-13B-v1.3} \citep{vicuna2023},
        \texttt{Vicuna-33B-v1.3} \citep{vicuna2023},
        \texttt{Vicuna-7B-v1.3} \citep{vicuna2023},
        \texttt{WizardLM-13B} \citep{xu2023wizardlm},
        \texttt{WizardLM-30B} \citep{xu2023wizardlm} \\
        \hline
    \end{tabular}
    \caption{List of models whose responses are used in Experiments \ref{7.2} and \ref{7.5}. Responses are pre-generated by MT-Bench except for \texttt{Llama-2-70B-chat-hf},
        \texttt{Falcon-180B-chat},
        \texttt{Openchat-3.5},
        \texttt{Mixtral-8x7B-Instruct-v0.1},
        \texttt{Mistral-7B-Instruct-v0.2},
        \texttt{Gemini-pro},
        \texttt{Dolphin-2.2.1-mistral-7B},
        \texttt{Solar-10.7B-instruct-v1.0}, and
        \texttt{Yi-34B-chat} in Experiment \ref{7.2}}
    \label{tab:modellist}
\end{table}



\clearpage
\subsection{Experiments \ref{7.3} and \ref{7.4} Additional Details}\label{ErrorBars7.3-7.4}

\begin{figure}[!htbp]
    \centering
    \includegraphics[width=.85\linewidth]{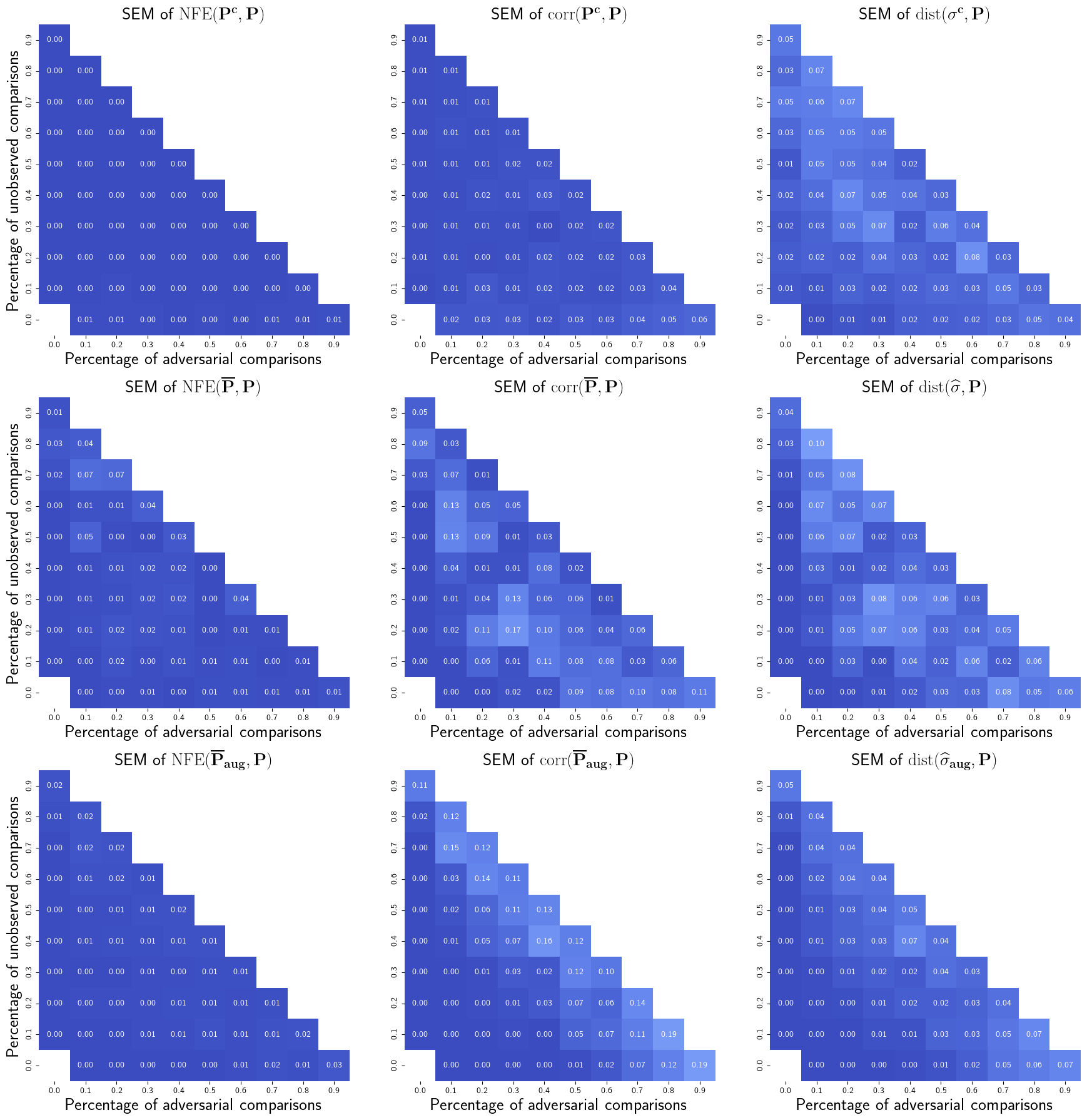}
    \caption{Standard Errors of the Mean of: Left column: NFE between reconstructed and original matrices. Middle column: Correlation between reconstructed and original matrices. Right column: Distance between reconstructed and original rankings. Top row: Unobserved and adversarial corruptions. Middle row: Recovering without augmentation. Bottom row: Recovering with augmentation.}
    \label{fig:enter-label}
\end{figure}

\clearpage
\section{Prompt template and answer of LLM judge}

We evaluate various responses from different Language Models (LLMs) using \texttt{GPT-4-1106-preview} \citep{openai2023gpt4} as the judge. We have adapted the prompt template from MT-Bench \citep{zheng2023judging}, but instead of scoring each candidate separately, we score all models in a single turn. Please note that the results generated by \texttt{GPT-4-1106-preview} judge may not be the same when the model gets updated. To ensure that the judge's response is deterministic, we have set the temperature parameter to 0.

\begin{figure}[h!]

\begin{tcolorbox}[colframe=RoyalBlue, colback=white]
{\scriptsize
\textbf{System prompt:} 
You are a helpful assistant.

\textbf{Prompt Template:} 
[Instruction]
Please act as an impartial judge and evaluate the qualities of the responses provided by many AI assistants to the user question displayed below. Your evaluation should consider factors such as the helpfulness, relevance, accuracy, depth, creativity, and level of detail of the response. Begin your evaluation by providing a short explanation to each of the assistant responses. Be as objective as possible. Your judgment should be based solely on the content of the answers, disregarding factors like the position of the responses, the length of the responses, or the names of the assistants. After providing your explanation, you must rate the responses on a scale of 0 to 1 by strictly following this format: 

Final verdict for each assistant (full list):

\{`Assistant A': $score_a$,

...
    
'Assistant n': $score_n$\}

For example:

Final verdict for each assistant (full list):

\{`Assistant A': 0.30,

`Assistant B': 0.19,

`Assistant C': 0.75\}

[The Start of Assistant A's Answer]

Assistant A's Answer

[The End of A's Answer]

...

[The Start of Assistant n's Answer]

Assistant n's Answer

[The End of n's Answer]
}

{\scriptsize
\textbf{Judge's Answer:} 
}

\begin{tcolorbox}[colframe=RoyalBlue, colback=gray!10]
{\scriptsize
Final verdict for each assistant (full list):

\begin{multicols}{3}
\{`Assistant A': 0.90,
    
`Assistant B': 0.95,
    
`Assistant C': 0.80,
    
`Assistant D': 0.75,
    
`Assistant E': 0.85,
\columnbreak
    
`Assistant F': 0.70,
    
`Assistant G': 0.60,
    
`Assistant H': 0.65,

`Assistant I': 0.55,

`Assistant J': 0.40,
\columnbreak
    
`Assistant K': 0.45,
    
`Assistant L': 0.50,
    
`Assistant M': 0.20,
    
`Assistant N': 0.35,
    
`Assistant O': 0.30\}
\end{multicols}    

Explanation of Scores:

A scored well as it provides a structured, detailed, and vivid account of the Hawaiian experience, including cultural and natural attractions.

B earned the highest score for a well-written, engaging narrative combining in-depth cultural experiences with comprehensive overviews of natural attractions, presenting it all in an engaging manner.

C's response was immersive, providing a daily itinerary format which gives readers a chronological perspective but loses some detail and depth in the descriptions, resulting in a slightly lower score.

D's response was engaging, offering personal reflections and detailed descriptions of the experiences, but it could capitalize on a more informative angle regarding cultural aspects.

E crafted an elaborate and expressive narrative with a strong focus on cultural elements and natural beauty. It's highly descriptive, which enhances its appeal.

F offered a nice overview with personal insights but was more generic compared to others, with less distinctive detail on the cultural elements and attractions.

G provided a good variety of information but lacked a personal touch and the engaging storytelling present in others, hence a middle-of-the-road score.

H put forth a decent attempt to cover cultural experiences and attractions, but the content was less rounded and not as rich in personal perspective.

I's storytelling lacked in-depth detail and read more like an overview, resulting in a lower rating compared to others that provided richer narratives.

J's response was brief and lacked the depth, detail, and storytelling elements found in the higher-scoring responses.

K focused more on personal experiences rather than providing informative cultural insights, which resulted in a score below others that balanced both elements better.

L offered an informative narrative but was less dynamic and less vivid in description than higher-scoring answers.

M’s response was too factual and lacked the storytelling and descriptive engagement present in other responses, leading to a lower score.

N provided a decent overview but lacked the richness of storytelling and description, leaving it towards the lower end of the scale.

O’s draft was informative but lacked the detail and narrative quality that made other responses stand out, resulting in a lower score.

}
\end{tcolorbox}

    \end{tcolorbox}
    \label{fig:gpt4}
\end{figure}


\clearpage
\section{Proofs}
\subsection{Proof of Claim~\ref{claim:claim1}}
\begin{proof}
{}Assume that we are exactly given the entries of the preference matrix as opposed to sampling them. Note that in order to estimate a ranking from a given preference matrix, we still need to use a pairwise ranking procedure. Let  $\se \in \S_n$ be the output of any Pairwise Ranking (PR) procedure with respect to an underlying preference matrix $\mathbf{Q} \in \P_n$. For a constant $\gamma > 1$, $\se$ is said to be $\gamma$-approximate if $\dist(\se,\mathbf{Q}) \leq \gamma \min_{\sigma \in \S_n} \dist(\sigma,\mathbf{Q})$. Define the following distance which measures the fraction of response pairs over which two preference matrices $\{ \mathbf{Q}, \mathbf{R} \} \in \P_n$ disagree.
\begin{small}
\begin{align*}
\dist \paran{\mathbf{Q}, \mathbf{R}} & := \begin{pmatrix}n \\ 2 \end{pmatrix}^{-1} \sum_{i<j} \mathbbm{1} \paran{(Q_{ij} > 1/2) \wedge (R_{ij} < 1/2)} \\
& + \begin{pmatrix}n \\ 2 \end{pmatrix}^{-1} \sum_{i<j} \mathbbm{1} \paran{(Q_{ij} < 1/2) \wedge (R_{ij} > 1/2)}
\end{align*}
\end{small}
By Lemma 20 of \citep{rajkumar2016can}, for $\mathbf{Q} \in \P_n^{ST}$ and $\mathbf{R} \in \P_n$, we have
$\dist(\se,\mathbf{Q}) \leq (1+\gamma) \dist(\mathbf{Q},\mathbf{R})$. But note that it is possible that $\dist \paran{\mathbf{Q}, \mathbf{R}} = 1$ as it is easy to construct by $\mathbf{R}$ that disagrees with $\mathbf{Q}$ in every entry by simply setting $\mathbf{R} = \mathbf{Q}^\top$.  
Now, we may set $\mathbf{Q} = \mathbf{P}$ and $\mathbf{R} = \mathbf{\pc}$ for any algorithm that uses $\mathbf{\pc}$ for ranking; specifically, for the adversary satisfying Assumption~\ref{ass:adv}, we can see by a direct counting argument that $\dist \paran{\mathbf{Q}, \mathbf{R}} \leq \frac{d(2n-1-d)}{n(n-1)}$ which proves the claim.
\end{proof}

\subsection{Proof of Lemma~\ref{lem:logit}}
\begin{proof}
{}Both follow by using the definition of the logit function that $\psi(a) = \log (a/(1-a))$ and using the property that $\log(ab) = \log(a) + \log(b)$.
\end{proof}

\subsection{Proof of Theorem~\ref{thm:main}}\label{pf:thm:main}

\begin{proof}
{}Let $\tp_{ij}$ be the empirical probability estimate of $P_{ij}$. Note that we compute  $\tp_{ij} = \frac{1}{K} \sum_{k=1}^K y_{ij}^k$ from the given pairwise comparison dataset, $\aleph = \{(i, j, \{ y_{ij}^k \})\}$. Now, $\mathbf{\pe} = \mathbf{\tp} + \mathbf{S}$. By Lemma~\ref{lem:logit}, we may write the adversarially corrupted empirical probability estimate as $\psi(\mathbf{\pe}) = \psi(\mathbf{\tp}) + \mathbf{\ts}$ where $\mathbf{\ts} = \psi(\mathbf{\tp} + \mathbf{S}) + \psi(1-\mathbf{\tp})$. We have $\psi(\mathbf{\tp}) = \psi(\mathbf{P}) + \mathbf{\tn}$ where $\mathbf{\tn} = \psi(\mathbf{\tp}) - \psi(\mathbf{P})$. Now, this noise, $\mathbf{\tn}$, is purely due to finite-sample effects which can be controlled (using concentration arguments given in the inequality $\xi_3$ below) by driving it down to as small a value as we want by ensuring large enough number of comparisons for each pair. Note that we input $\psi(\mathbf{\pe}) = \psi(\mathbf{P})+\mathbf{\ts}+\mathbf{\tn}$ to Subroutine~\ref{alg:rpca} and obtain $\psi(\mathbf{\pec})$ as the output in Step 3 of Algorithm~\ref{alg:rpr}. Hence, using Theorem 2 from \citep{netrapalli2014non}, if $\infnorm{\mathbf{\tn}} \leq \sigma_{\min}(\psi(\mathbf{P}))/100n$, we have,
\begin{align*}
\frobnorm{\psi(\mathbf{\pec}) - \psi(\mathbf{P})} \leq \epsilon' + 2 \mu^2 r \paran{7 \twonorm{\mathbf{\tn}} + \frac{8n}{r} \infnorm{\mathbf{\tn}}}
\end{align*}
after $T \geq 10 \log (3 \mu^2 r \sigma_1 / \epsilon')$ iterations associated with Step 1 of Subroutine~
\ref{alg:rpca}. Next, we have, with probability at least $1-1/n^3$,
\begin{align*}
\frobnorm{\psi(\mathbf{\pec}) - \psi(\mathbf{P})} & \leq \epsilon' + 2 \mu^2 r \paran{7 \twonorm{\mathbf{\tn}} + \frac{8n}{r} \infnorm{\mathbf{\tn}}} \\
& \stackrel{\xi_1}{\leq} \epsilon' + 32 \mu^2 n \twonorm{\mathbf{\tn}} \stackrel{\xi_2}{\leq} \epsilon' + 32 \mu^2 n \tau \\
& \stackrel{\xi_3}{\leq} n \sqrt{\frac{\epsilon}{1+\gamma}} \frac{\Delta}{2}
\end{align*}
where $\xi_1$ follows by using $r \leq n$ and $\infnorm{\mathbf{\tn}} \leq \twonorm{\mathbf{\tn}}$, $\xi_2$ follows by substituting for $\mathbf{\tn}$ from Lemma~\ref{lem:sampling_noise}, and $\xi_3$ is obtained using $\epsilon' = n \sqrt{\frac{\epsilon}{1+\gamma}} \frac{\Delta}{4} $, $\tau = \min \paran{\sigma_{\min}(\psi(\mathbf{P}))/100, \sqrt{\frac{\epsilon}{1+\gamma}} \frac{\Delta}{128 \mu^2} }$. Then using similar arguments as proof of Theorem 13 in \citep{rajkumar2016can}, we obtain our result.
\end{proof}

\subsection{Lemma \ref{lem:sampling_noise}}
\begin{lemma}[\textbf{Concentration of Sampling Noise}]
\label{lem:sampling_noise}
Under the conditions 1-4 of Theorem~\ref{thm:main}, consider each response pair compared such that the number of comparisons per response pair \( K \) satisfies:

\begin{enumerate}[label=(3\alph*)]
    \item \label{itm:standard_case} $K \geq \frac{L^2 \log^2(n)}{\tau^2} \left( 2\log(2) + 10\log(n) \right)$ when the sampling noise matrix satisfies the inequality $\twonorm{\mathbf{\tn}} \leq 2\log{(n)} \infnorm{\mathbf{\tn}}$; then, with probability at least \( 1 - \frac{1}{n^3} \), \( \twonorm{\mathbf{\tn}} \leq \tau \).
    \item \label{itm:worst_case} $K \geq \frac{L^2 n^2}{2\tau^2} \left(5\log(n) + \log(2)\right)$ when the sampling noise matrix satisfies the inequality $\twonorm{\mathbf{\tn}} \leq n \infnorm{\mathbf{\tn}}$; then, with probability at least \( 1 - \frac{1}{n^3} \), \( \twonorm{\mathbf{\tn}} \leq \tau \).
\end{enumerate}

\end{lemma}

\subsubsection{Proof of Lemma \ref{itm:standard_case}}
\begin{small}
\begin{proof}
{}Let $L$ be the Lipschitz constant of $\psi$ and set $K \geq \frac{L^2 \log^2(n)}{\tau^2} \left( 2\log(2) + 10\log(n) \right)$. Using the inequality that $\twonorm{\mathbf{\tn}} \leq 2\log(n) \infnorm{\mathbf{\tn}}$,
\begin{align*}
\Pr \paran{ \twonorm{\mathbf{\tn}} \geq \tau} & \leq \Pr \paran{ \infnorm{\mathbf{\tn}} \geq \frac{\tau}{2\log{n}} } \\
& = \Pr \paran{ \exists (i,j) : \abs{ \psi(\pe_{ij}) - \psi(P_{ij}) } \geq \frac{\tau}{2\log{n}} } \\
& \leq  \sum_{i,j} \Pr \paran{ \abs{ \psi(\pe_{ij}) - \psi(P_{ij}) }  \geq \frac{\tau}{2\log{n}} } \\
& \leq  \sum_{i,j} \Pr \paran{ \abs{ \pe_{ij} - P_{ij} }  \geq \frac{\tau}{2\log{n} L} } \leq \frac{1}{n^3}
\end{align*}
\end{proof}
\end{small}

\subsubsection{Proof of Lemma \ref{itm:worst_case}}
\begin{small}
\begin{proof}
{}Let $L$ be the Lipschitz constant of $\psi$ and set $K \geq \frac{L^2 n^2}{2\tau^2} \left(5\log(n) + \log(2)\right)$. Using the inequality that $\twonorm{\mathbf{\tn}} \leq n \infnorm{\mathbf{\tn}}$,
\begin{align*}
\Pr \paran{ \twonorm{\mathbf{\tn}} \geq \tau} & \leq \Pr \paran{ \infnorm{\mathbf{\tn}} \geq \frac{\tau}{n} } \\
& = \Pr \paran{ \exists (i,j) : \abs{ \psi(\pe_{ij}) - \psi(P_{ij}) } \geq \frac{\tau}{n} } \\
& \leq  \sum_{i,j} \Pr \paran{ \abs{ \psi(\pe_{ij}) - \psi(P_{ij}) }  \geq \frac{\tau}{n} } \\
& \leq  \sum_{i,j} \Pr \paran{ \abs{ \pe_{ij} - P_{ij} }  \geq \frac{\tau}{n L} } \leq \frac{1}{n^3}
\end{align*}
\end{proof}
\end{small}

\subsection{Proof of Theorem~\ref{thm:btl}}

\begin{proof}
{}From Lemma~\ref{lem:incoh}, we have $\psi(\mathbf{P}) = \mathbf{1 w}^\top - \mathbf{w 1}^\top$ for the BTL model where $\psi$ is the logit function. Clearly, in this case, $\psi(\mathbf{P})$ is a real skew-symmetric matrix of rank $r=2$. Since it is skew-symmetric, its eigenvalues, which are the roots of its characteristic polynomial, are of the form $\pm \lambda i$ for some $\lambda \in \R$ and $i = \sqrt{-1}$, and hence, $\sigma_{\min}(\psi(\mathbf{P})) = \sigma_{\max}(\psi(\mathbf{P}))$, ie, the condition number of $\psi(\mathbf{P})$, $\kappa = 1$. Now, we recall the spectral-lower bound from Corollary 2 of \citep{horne1997lower},
\begin{equation}
\label{eqn:lb}
\sigma_{\min}(\psi(\mathbf{P})) \geq \frac{\frobnorm{\psi(\mathbf{P})}}{\sqrt{r(r-1)}} \geq \sqrt{\frac{n(n-1) }{2}} \Delta_w 
\end{equation}
where $\Delta_w = \min_{i,j} \abs{w_i-w_j}$.

Let $\Omega \subseteq [n]\times[n]$ be a subset of all the response pairs with comparison results among which some might be corrupted by sparse noise, ie, $\psi(\mathbf{\pe}_{\Omega}) = \psi(\mathbf{P}_{\Omega})+\mathbf{\ts}_{\Omega}+\mathbf{\tn}_{\Omega}$. Let $\mathbf{T} := \mathbf{\ts}_{\Omega}+\mathbf{\tn}_{\Omega}$. From Theorem 1.2 of \citep{keshavan2010matrix}, we have $\frac{1}{n} \frobnorm{\psi(\mathbf{\pe}) - \psi(\mathbf{P})} = \frac{1}{n} \frobnorm{\mathbf{T} + \mathbf{M}} \leq C \kappa^2 \frac{n \sqrt{r}}{\abs{\Omega}} \twonorm{\mathbf{T}}$ where $\mathbf{M}$ is the noise matrix after obtaining the completed matrix $\psi(\mathbf{\pe})$ from $\psi(\mathbf{\pe}_{\Omega})$ using OptSpace. 

Using triangle inequality and noting that $\abs{\Omega} \geq C'' n \log(n)$, the noise may be bounded as
\begin{small}
\begin{align}
\infnorm{\mathbf{\tn}_\Omega + \mathbf{M}} & \leq \frobnorm{\mathbf{\tn}_\Omega + \mathbf{M}} \leq \twonorm{\mathbf{T}} \frac{\sqrt{2} C n^2}{\abs{\Omega}} + \frobnorm{\mathbf{\ts}_\Omega} \nn \\
& \stackrel{\zeta_1}{\leq} C' \frac{n}{\log(n)} \twonorm{\mathbf{\ts}_\Omega}
\label{eqn:ub}
\end{align}
\end{small}
where $C$, $C'$ and $C''$ are constants and $\zeta_1$ is obtained by using the triangle inequality that $\twonorm{\mathbf{T}} \leq \twonorm{\mathbf{\ts}_\Omega} + \twonorm{\mathbf{\tn}_\Omega}$, followed by setting $K \geq c n^4 / \Delta_w$ for constant $c$ and finally using $\frobnorm{\mathbf{\ts}_\Omega} \leq \sqrt{n} \twonorm{\mathbf{\ts}_\Omega}$. 

Then, combining Equations \ref{eqn:ub} and \ref{eqn:lb}, we have if
\begin{small}
\begin{align*}
\frac{\log(n)}{C_\Delta n} \Delta_w & \geq \twonorm{\mathbf{\ts}_{\Omega} } = \twonorm{\psi(\mathbf{\pe}) - \psi(\mathbf{\tp})} \\
& \geq \infnorm{\psi(\mathbf{\pe}) - \psi(\mathbf{\tp})} \geq
L \infnorm{\mathbf{\pe} - \mathbf{\tp}} \geq \infnorm{\mathbf{S}}
\end{align*}
\end{small}
where $C_\Delta$ is a global constant and using Lemma~\ref{lem:sampling_noise}, then we have the guarantee along similar lines as that of Theorem~\ref{thm:main} that Algorithm~\ref{alg:rpopr} returns an estimated permutation which satisfies $\dist(\se,\mathbf{P}) \leq \epsilon$.
\end{proof}